\documentclass[10pt]{article} 

\usepackage[preprint]{rlj} 

\usepackage{amssymb}   
\usepackage{mathtools}   
\usepackage{mathrsfs}    
\usepackage{graphicx}      
\usepackage{subcaption}   
\usepackage[space]{grffile}   
\usepackage{url}
\usepackage{lipsum} 
\usepackage{floatflt}

\usepackage{amsthm} 
\usepackage{amsmath} 
\usepackage{bbm}
\usepackage{booktabs}
\usepackage{multirow}

\usepackage{algorithm2e}
\RestyleAlgo{ruled}
\SetKwComment{Comment}{/* }{ */}
\usepackage{algpseudocode}

\newcommand{\R}{\ensuremath{\mathbb{R}}}   
\newcommand{\N}{\ensuremath{\mathbb{N}}}   

\def\A{{\cal A}}
\def\S{{\cal S}}

\newtheorem{lemma*}{Lemma}
\theoremstyle{definition}

\definecolor{deepred}{rgb}{0.75, 0.2, 0.2}
\definecolor{forestgreen}{rgb}{0.13, 0.55, 0.13}
\definecolor{royalpurple}{rgb}{0.55, 0.2, 0.75}

\title{Chargax: A JAX Accelerated EV Charging Simulator}
\setrunningtitle{Chargax: A JAX Accelerated EV Charging Simulator}

\author{Koen Ponse \textsuperscript{1, $\dagger$}, Jan Felix Kleuker\textsuperscript{1, $\dagger$}, Aske Plaat\textsuperscript{1}, Thomas Moerland\textsuperscript{1}}

\emails{k.ponse@liacs.leidenuniv.nl}
\affiliations{
$^\dagger$ Equal Contribution,
$^{1}$\textbf{Leiden Institute of Advanced Computer Science, Leiden University, the Netherlands}}

\contribution{
    This paper presents Chargax, an open-source EV charging environment written in JAX
}{ Chargax could be used as a high-performance test bed for reinforcement learning benchmarking, or to develop better control algorithms for EV charging. }
\contribution{
    Comparisons in performance are made with previously existing EV simulators for RL that demonstrate Chargax decreases training times by a factor of 100x or more.
} {Prior work established EV charging simulators for RL that did no leverage the GPU}
\contribution{
    We perform additional experiments validating reinforcement learning training in a variety of scenarios, data distributions shifts, and reward objectives.
}{ None }
\contribution{
    We create an explicit split in the state space which highlights the interchangeable parts in the Chargax environment. This modularity allows representation of diverse real-world charging station configurations and scenarios.
} {Prior work often used this split implicitly, and allow for less customisability}


\summary{
    Deep Reinforcement Learning can play a key role in addressing sustainable energy challenges. For instance, many grid systems are heavily congested, highlighting the urgent need to enhance operational efficiency. However, reinforcement learning approaches have traditionally been slow due to the high sample complexity and expensive simulation requirements.
    While recent works have effectively used GPUs to accelerate data generation by converting environments to JAX, these works have largely focussed on classical toy problems.
    This paper introduces Chargax, a JAX-based environment for realistic simulation of electric vehicle charging stations designed for accelerated training of RL agents. 
    We validate our environment in a variety of scenarios based on real data, comparing reinforcement learning agents against baselines.
    Chargax delivers substantial computational performance improvements of over 100x-1000x over existing environments. Additionally, Chargax' modular architecture enables the representation of diverse real-world charging station configurations.

}

\begin{document}

\maketitle  

\begin{abstract}

    Deep Reinforcement Learning can play a key role in addressing sustainable energy challenges. For instance, many grid systems are heavily congested, highlighting the urgent need to enhance operational efficiency. However, reinforcement learning approaches have traditionally been slow due to the high sample complexity and expensive simulation requirements.
    While recent works have effectively used GPUs to accelerate data generation by converting environments to JAX, these works have largely focussed on classical toy problems.
    This paper introduces Chargax, a JAX-based environment for realistic simulation of electric vehicle charging stations designed for accelerated training of RL agents. 
    We validate our environment in a variety of scenarios based on real data, comparing reinforcement learning agents against baselines.
    Chargax delivers substantial computational performance improvements of over 100x-1000x over existing environments. Additionally, Chargax' modular architecture enables the representation of diverse real-world charging station configurations.\footnote{\label{footnote:sourcecode}Available on GitHub at \url{https://github.com/ponseko/chargax}}

\end{abstract}

\section{Introduction}

\begin{floatingfigure}[r]{0.49\linewidth}
    \centering
    \includegraphics[trim=3 10 3 10,clip,width=0.47\textwidth]{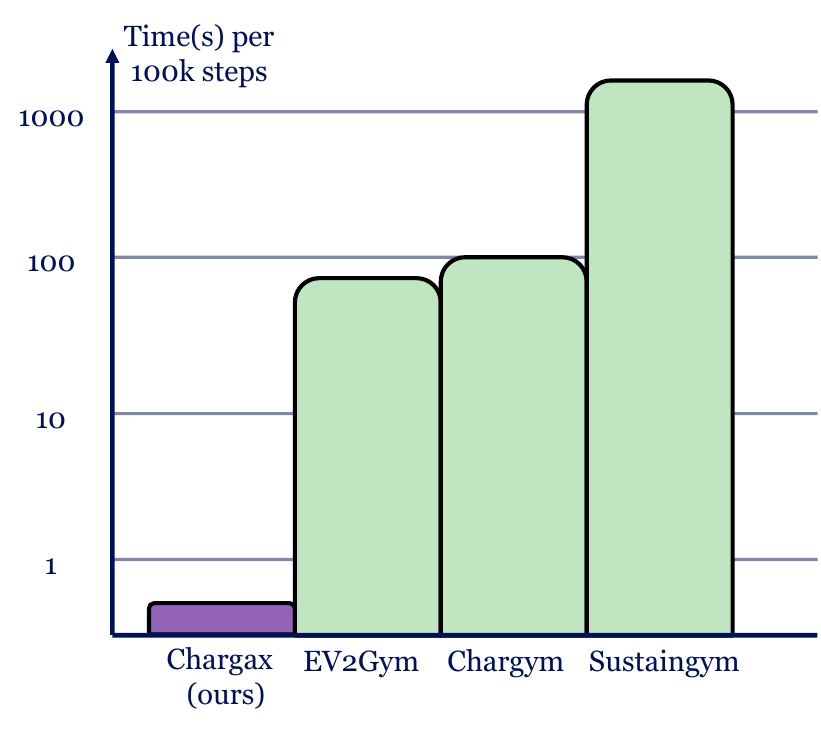}
    \caption{Comparison between Chargax and prior EV Gym Environments in seconds to complete 100k training steps using PPO. See Table \ref{tab:performance} for a more complete overview.}
    \label{fig:speedup}
\end{floatingfigure}

Deep Reinforcement Learning (RL) can approximate optimal policies for difficult decision problems that are impossible to solve with traditional mathematical methods.
Such problems occur frequently in sustainable energy challenges such as operation of windfarms \citep{fernandez-gaunaActorcriticContinuousState2022}, electric vehicle charging \citep{rehman2024grid}, and nuclear fusion reactors \citep{seo2024avoiding}. While RL has achieved successful solutions to these challenges, further development of RL algorithms hinges on the availability of realistic simulation environments and benchmarks \citep{ponse2024reinforcement}.

Unfortunately, reinforcement learning is notoriously sample-inefficient \citep{yarats_improving_2020, kaiser_model-based_2024}. It often requires many environments samples which are slow and possibly expensive to generate. These simulations have often been running on the CPU -- disallowing RL researchers from truly harvesting the potential scale-up of GPUs that other machine learning fields have been enjoying \citep{MLST}. To this end, the development of RL environments using JAX \citep{jax2018github} has recently gained increasing attention \citep{brax2021github,gymnax2022github,pignatelli2024navix,bonnet2024jumanji}.

However, current implementations remain largely confined to simplified toy problems, highlighting a significant gap in real-world applications utilizing JAX.

\begin{figure}[ht]
    \centering
    \includegraphics[width=0.90\linewidth]{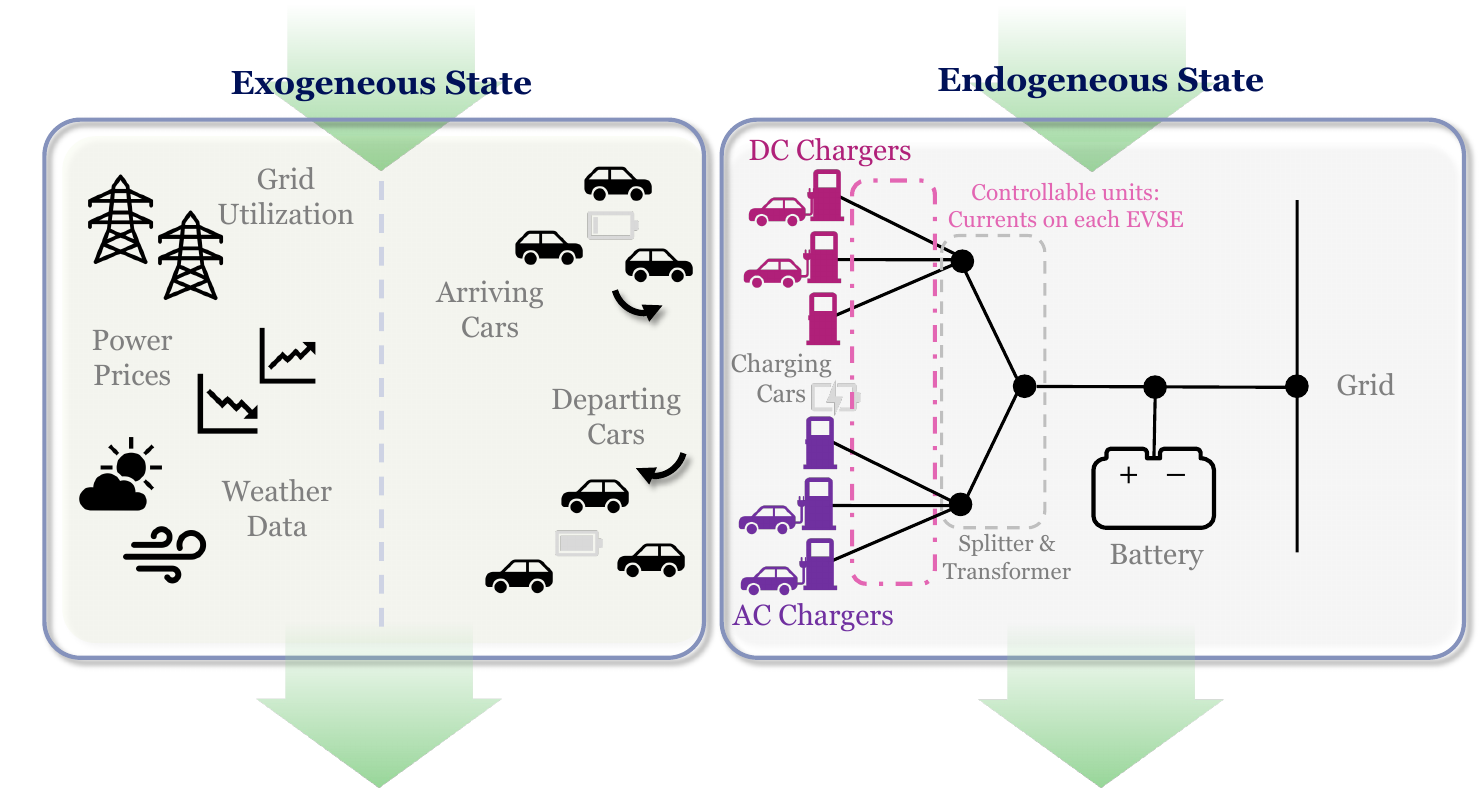}
    \caption{An overview of the Chargax environment. The \textit{endogenous state} describes the state variables that are influenced directly by the agent. The \textit{exogenous state} evolves via, agent-independent, predefined time series data.}
    \label{fig:01}
\end{figure}

\paragraph*{Contribution} In this work, we aim to bridge this gap by introducing, to the best of our knowledge, the first reinforcement learning environment for EV charging implemented in JAX.
\begin{itemize}
    \item Our environment, \textit{Chargax}, achieves a significant speedup of 100x-1000x compared to existing environments for EV charging \citep{yehSust, orfanoudakEV2G2024, karatzinis2022chargym}. This lowers training times from hours or even days to mere minutes -- allowing for orders of magnitude more experiments (see Figure \ref{fig:speedup}). 
    \item Chargax extends the generalisability of existing frameworks. As highlighted in a recent survey \citep{alaeeRevi2023}, optimising electric vehicle charging strategies involves a diverse set of potential objectives. We demonstrate that many of these objectives can be addressed within a single simulation framework by ensuring sufficient flexibility.
    \item Chargax can function as a high-performance test bed for reinforcement learning benchmarking on real-world applications. Empirically, we demonstrate how RL agents are able to outperform baselines and allow for flexible goals such as user satisfaction. We open source \textit{Chargax}$^{\ref{footnote:sourcecode}}$ for the wider community to experiment with.
\end{itemize}

\textit{Chargax} is equipped with predefined datasets, reward functions, and charging station architectures for various scenarios. Moreover, all components are fully customizable, enabling researchers to tailor the environment to specific requirements, thereby facilitating efficient and adaptable RL-based solutions for EV charging optimization.

\section{Related Work}

Prior work in EV charging includes the gym environments Sustaingym \citep{yehSust} (based on \citep{leeACNS2020}), Chargym \citep{karatzinis2022chargym}, and the more recently released EV2Gym \citep{orfanoudakEV2G2024}. Compared to the works of \citealt{yehSust, leeACNS2020} and \citealt{karatzinis2022chargym} our framework provides additional flexibility for the architecture of the charging station, scenario selection, and the customer and car profiles. Compared to \citealt{orfanoudakEV2G2024}, which also prioritises flexibility, our approach features a more streamlined state and architecture representation. To the best of our knowledge, Chargax is the only
Gym-like environment that includes local car and price data across multiple regions. Furthermore, Chargax is orders of magnitude faster and in turn allows for large scale experiments on the GPU (See Figure \ref{fig:speedup}). Apart from these Gym-like simulators, there exist a history of EV charging simulators \citep{saxena2013v2g, rigas2018evlibsim, balogun2023ev, canigueral2023evsim}.

In recent years, many classical Gym environments have been reimplemented in JAX. We direct the reader to the following non-exhaustive list \citep{brax2021github,gymnax2022github, nikulin2023xlandminigrid,flair2023jaxmarl,koyamada2023pgx,pignatelli2024navix,bonnet2024jumanji}. These implementations have largely been reimplementations of classical toy problems, highlighting a gap in environments modelling real-world problems.

\section{Preliminaries}

\subsection*{Markov Decision Process}

Formally, an environment is represented as a Markov Decision Process (MDP; \citealt{suttonRein2018}) defined by a tuple $\mathcal{M} = (\S,\A,p_0,p,r,\gamma)$. Here, $\S$ is a state space, $\A$ is a action space, $p_0 \in \Delta(\S)$\footnote{$\Delta(\mathcal{X})$ denotes the set of probability distributions over a set $\mathcal{X}$} is the initial state distribution, $p(\cdot|s,a) \in \Delta(\S)$ is the probabilistic transition function, $r(s,a, s')$ denotes the reward function and $\gamma \in [0,1)$ is the discount factor. In the next section (\ref{sec:environment-design}), we provide a detailed discussion of the motivation behind the choices for each MDP component and formally define these quantities used within the framework.

\subsection*{JAX}

JAX is a Python library aimed at accelerator-orientated programming with a NumPy interface \citep{jax2018github}. It offers function transformations to perform, for example, just-in-time-compilation, vectorization, and differentiation. Although JAX is a common foundation for deep learning frameworks \citep{flax2020github, kidger2021equinox}, its just-in-time compilation transformation allows users to easily run plain Python code on accelerators such as GPUs and TPUs. Although JAX imposes some constraints on how these functions should be constructed, it enables complete environment transition functions to operate on the GPU. This allows many more operations and environments to run in parallel and eliminates data transfers between the CPU and GPU for gradient descent updates, both of which can potentially decrease the computational time requirements of reinforcement learning experiments significantly~\citep{lu_luchris429purejaxrl_2024, hessel2021podracer}.

\section{Environment Design} \label{sec:environment-design}

In many real-world control environments not all state variables are directly affected by the actions of the agent. Instead, some of the state variables transition into their next state via an (agent-independent) function (often time series). These functions often rely on some external data source and therefore these variables describe exactly the entry points for data integration that can be flexibly interchanged within Chargax. Although this data distinction is often implicitly present \citep{ponse2024reinforcement}, we will formalise this separation explicitly in Chargax to make clear which parts of the state can flexibly be interchanged.

Consequently, we split the environment state in an \textit{endogenous} and an \textit{exogenous} state space. 
The endogenous state space refers to the typical state variables that are influenced by the agents' actions during the transition function.
In contrast, exogenous state variables transition into their next state via an (agent-independent) time series. Examples of exogenous state variables are weather variables, or national electricity prices. Even though these variables are not affected by the agents' actions, they may influence the agent by providing an additional learning signal and/or alter the reward. 

An overview of Chargax is shown in Figure \ref{fig:01} and in the following we provide a high-level overview of the Chargax environment. Full implementation details, including all equations for transition dynamics and reward functions, are provided in Appendix \ref{ap:environment-details}.

\subsection*{EV Station Layout}

\begin{figure}[t]
    \centering
    \includegraphics[width=0.95\linewidth]{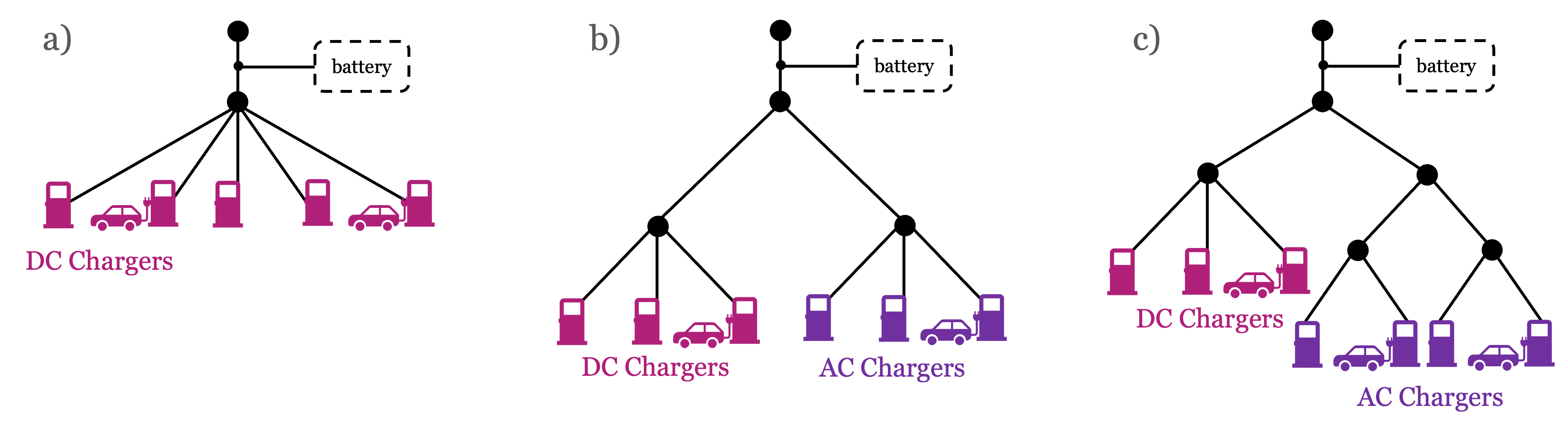}
    \caption{Trees representing different architectures: a) simplest scenario, one type of charger; b) multiple types of chargers, one splitter per charging type; c) multiple types of chargers with multiple splitters per type, imposing additional constraints on the currents. Each node represents a combination of splitters, transformers, cables, and other electrical components.
    }
    \label{fig:charge-tree}
\end{figure}

When initialising a Chargax environment, a fixed architectural design for the station is generated or provided. This design is fixed and, therefore, not influenced by the transition function. We represent this electronic infrastructure of the charging station in the form of a tree \citep{leeAdap2021}, with leaves representing the charging ports (Electric Vehicle Supply Equipment; EVSE; \citet{leeACNS2020}) (see Figure \ref{fig:charge-tree}). The root node represents the grid connection access, and all other nodes represent a combination of splitters, cables, and transformers, and are equipped with a maximum power capacity and efficiency coefficient, imposing constraints on the system. In Chargax, we additionally assume a fixed voltage $V$ for each of the EVSEs in the architecture.

Chargax supplies methods for generating some charging station architectures. However, custom architectures can be built by constructing a tree of simple nodes to mirror existing real-world infrastructure.

\subsection*{Endogenous State Space}

The endogenous state consists of the state of the various charging ports and their connected cars, and the station battery. 
As each charging port (and the battery) has a fixed voltage level, we allow the actual power drawn to be regulated by controlling the current \citep{orfanoudakEV2G2024}. Losses are incorporated through efficiency coefficients at each node (including the charging ports).

In addition to the set current at each charging port ($I_{\mathrm{drawn}}(t) \in [0, I_\mathrm{max}]$), and whether the port is currently occupied ($\mathbbm{1}_\mathrm{occup}$), the endogenous state contains information for the connected cars. This includes their state-of-charge (SoC) and the remaining required power $\Delta E_\mathrm{remain}$. Additional information for each car is supplied exogenously and remains fixed until the car leaves. We will expand on this information in the next section.
The endogenous state space can optionally be expanded with a station battery. This battery is modelled similarly to an EVSE -- with a fixed voltage and controlled via the set current. The battery allows the agent to store energy to facilitate effective discharging strategies. 
In brief summary, the endogenous state is represented by:
\begin{itemize}
    \item For each EVSE: $I_{\mathrm{drawn}}(t)\in \R_{\geq 0}$, $\mathbbm{1}_\mathrm{occup}(t) \in \{0,1\}$, $\Delta E_\mathrm{remain}$, SoC$(t)$
    \item Battery: $I_\mathrm{battery}(t), \mathrm{SoC}_\mathrm{battery}(t)$
\end{itemize}

Enumerating the existing EVSEs by $i=1,\dots,N$, the total endogenous state space can be expressed as $s_\mathrm{end} = (s_\mathrm{battery}, s_{c,1},\dots,s_{c,N})$. A complete overview of the state space is given in Appendix~\ref{ap:state-space}.

\begin{table}[b]
\centering
\caption{Overview of available Profiles in Chargax. Default settings are marked in bold.}
\begin{tabular}{l|l|l|l|l}
Price Profiles  & Architectures                            & Car Distributions & Arrival Frequency   & User Profiles       \\ \hline
\textbf{NL} &  Simple: Single                  & \textbf{Europe}   & Low Traffic     & Highway             \\
FR              &      Charger Type & US                & \textbf{Medium Traffic}  & Residential \\
DE           &          \textbf{Simple: Multiple}                  & World             & High Traffic    & Work             \\
\textit{Custom} &      \textbf{      Charger Types}                                     & \textit{Custom}   & \textit{Custom} & \textbf{Shopping}       \\
                &      \textit{Custom}                                    &                   &                 & \textit{Custom}    
\end{tabular}
\label{table:data-overview}
\end{table}

\subsection*{Exogenous State Space}

As described previously, the exogenous state variables evolve independently of the agent's actions. As such, the remainder of the variables discussed here are typically sampled from distributions that are generated via a provided time series or some predefined function. Currently, Chargax works with exogenous state variables for arrival data, user profiles, car profiles, and grid price data.

The \textbf{arrival data} represents the number of cars that arrive at a given timestep. Typically, this depends on the time and location of the charging station. Likewise, the location can also stipulate the typical \textbf{user profile} of the arriving cars. This profile describes the state of the car that is induced by their owner, such as the arrival SoC, desired charging level, and time of departure. 
\textbf{Car profile} variables are derived from the physical properties of the cars themselves. These include the maximum capacity of the car battery and the maximum charge speed.
Lastly, the \textbf{grid prices} are an important exogenous variable for calculating the profit, which is often a large factor in the reward.

Chargax comes equipped with a variety of standard datasets (see Table \ref{table:data-overview}), most of which are based on real data. These datasets can be used to sample exogenous variables that resemble realistic scenarios. For example, Europe and the US have a different distribution of electric vehicles on the road; in turn, the distribution of charging demands is different in both regions. While datasets are provided, Chargax is built such that users can use their own data or functions to populate these variables.

\subsection*{Action Space}

At each timestep, the agent controlling the charging station can adjust the power at each EVSE by altering the current \citep{orfanoudakEV2G2024}, 
i.e. an action is characterized as $$a(t) = (\Delta I_i(t))_{i=1}^{N+1} \in \R^{N+1}.$$

Here, for the sake of notational convenience, the battery is treated as the $N+1$-th charging pole. Notably, the agent cannot accept/decline cars and is assumed to serve arriving cars, as long as there are free spots.

\subsection*{Transition Function}

At a high level, the transition function consists  of four sequential steps, which we detail below. Full implementation details can be found in Appendix \ref{ap:transition-function}.

\begin{itemize}
    \item \textbf{Apply Actions} First, we apply the agent's to adjust the power drawn by each charging port. We limit the maximum power by the capacity of the port, as well as the current maximum (dis)charging rate of the car stationed at each charging port.
    
    \item  \textbf{Charge Stationed Cars} With the newly set power levels, we (dis)charge each car over the time interval of a timestep. Here, we assume a constant charging rate over the full interval $\Delta t$.
    
    \item  \textbf{Departure of Cars} Next, cars fully charged (charge-sensitive users) or with no time remaining (time-sensitive users) will leave.
    
    \item  \textbf{Arrival of new Cars} Finally, an amount of new cars will be sampled through our exogenous data, along with a \textit{user profile} and \textit{car profile}. The amount of new cars is clipped by the number of free spots available and the remaining cars are automatically rejected. 
    Arriving cars will park in the first available spot as provided by the provided station architecture.
\end{itemize}

\subsection*{Reward Function}

In RL, the reward functions reflects the notion of optimality, i.e. the desired behaviour. 
In this section, we outline some of the reward functions that are available in Chargax, and how they reflect different objectives. We provide additional details in Appendix \ref{ap:rewards}.

\paragraph*{Profit Maximisation} Profit maximisation  lies at the core of most Charging Station Operations \citep{alinejadOpti2021, changCoor2021, mirzaeiTwos2021a, yeLear2022}. The amount of net energy transferred into cars in the interval $[t,t+\Delta t]$ is denoted by $\Delta E_\mathrm{net}(t)$. The amount of energy fed into the grid as a result from discharging cars is denoted by $\Delta E_{\to \mathrm{grid}}(t)$, and the amount of energy that has to be drawn from the net to transfer set levels of energy into cars $\Delta E_{\mathrm{grid} \to}(t)$. Lastly, the energy contributed by (dis-)charging the battery $\Delta E_\mathrm{b, net}(t)$ has to be incorporated, resulting in the following net energy that is drawn from (or pushed into) the grid
\begin{equation}
    \Delta E_{\mathrm{grid, net}} = \Delta E_{\mathrm{grid} \to}(t) + \Delta E_{\to \mathrm{grid}}(t) + \Delta E_\mathrm{b, net}(t).
\end{equation}
We further assume that the price at which we sell and buy power from car owners is the same, i.e. $p_\mathrm{sell}$. This results in the following profit
\begin{equation}
    \Pi(t) = \begin{cases}
        p_\mathrm{sell}(t) \cdot \Delta E_\mathrm{net}(t) - p_\mathrm{buy}(t) \cdot \Delta E_{\mathrm{grid, net}} - c_{\Delta t}& \Delta E_{\mathrm{grid, net}} > 0, \\
        p_\mathrm{sell}(t) \cdot \Delta E_\mathrm{net}(t) - p_\mathrm{sell, grid}(t) \cdot \Delta E_{\mathrm{grid, net}} - c_{\Delta t} & \Delta E_{\mathrm{grid, net}} \leq 0.
    \end{cases}
\end{equation}
Here, $c_{\Delta t}$ denotes the fixed cost for running the facility per $\Delta t$.

\paragraph*{Profit Maximisation under constraints} To further steer agents' learnt behaviour in a direction, constraints can be induced to penalise certain (undesired) behaviour through penalty terms $c(t)$. The resulting reward will be the profit minus the linear combination of (possibly) multiple penalty terms
\begin{equation}  \label{eq:alpha-reward}
    r(t) = \Pi(t) - \sum_{c} \alpha_c \,c(t). 
\end{equation}

Different linear combinations of different penalty terms allow Chargax to be flexible in its optimization objective. Chargax comes equipped with various of these penalty terms to better optimize for, for example, customer satisfaction, battery degradation, or violating node constraints. We provide a more complete list of possible penalty terms along with a formal expression in Appendix \ref{ap:rewards}. However, we emphasise that these are mere suggestions, and that these rewards are not comprehensive in reflecting the full landscape of Charging Station Optimisation challenges, and we encourage users to customise their reward function within the provided framework.

\section{Experiments} \label{sec:experiments}

In this section, we demonstrate the use of Chargax across different included scenarios. Additionally, we highlight performance improvements of Chargax compared to previous EV charging simulations. Full details of the used model and configuration parameters, along with additional experimental results, can be found in Appendix \ref{sec:AppendixImplementationDetails} and \ref{sec:AppendixAdditionalExperiments} respectively.

\begin{figure}[t]
    \centering
    \begin{subfigure}[b]{0.32\textwidth} 
        \centering
        \includegraphics[width=\textwidth]{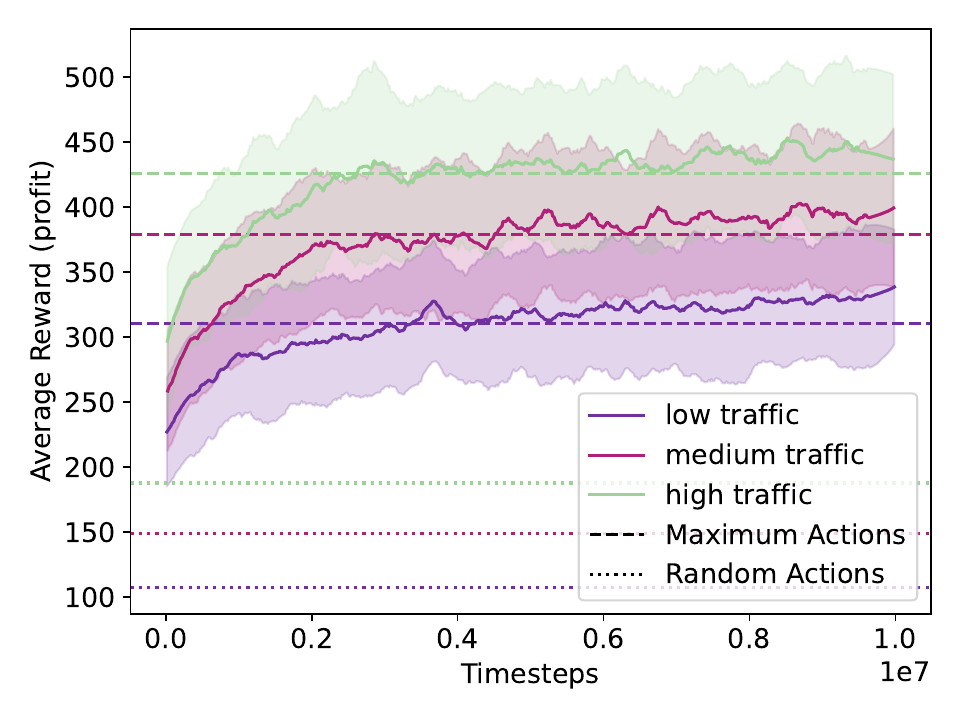}
        \caption{PPO in Shopping scenario} \label{fig:learning_curve}
    \end{subfigure}
    \hfill
    \begin{subfigure}[b]{0.32\textwidth} 
        \centering
        \includegraphics[width=\textwidth]{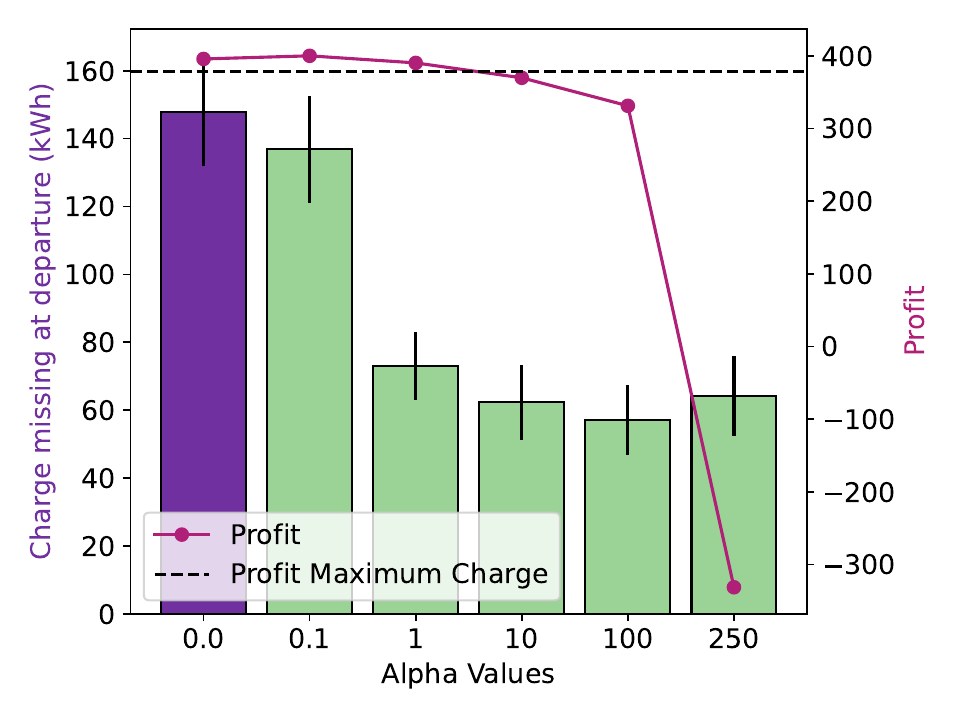}
        \caption{Charge missing at departure} \label{fig:alpha_ablation_a}
    \end{subfigure}
    \hfill
    \begin{subfigure}[b]{0.32\textwidth}
        \centering
        \includegraphics[width=\textwidth]{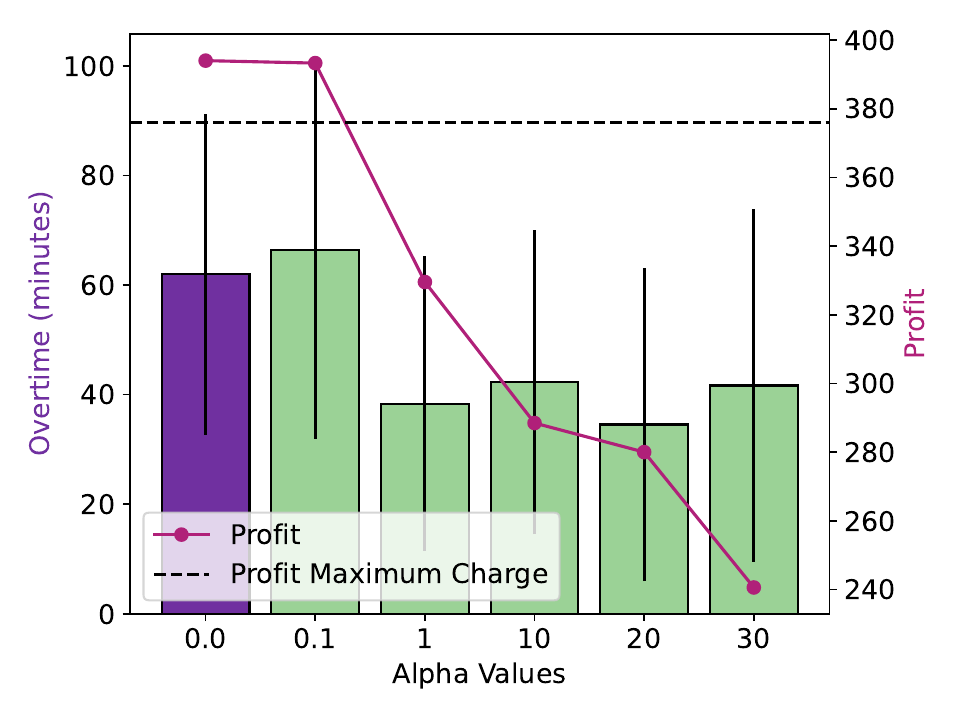}
        \caption{Charged overtime} \label{fig:alpha_ablation_b}
    \end{subfigure}
    \caption{In a) average episode rewards during training a PPO agent in the shopping scenario with different levels of traffic. The RL solution manages to increase profit over the baseline that always charged the maximum possible amount. In b) and c), user satisfaction measured as charge (kWh) missing at time of departure (b), and time exceeded to fully charge cars (c). Higher $\alpha$-values weigh the measured variable greater in the reward (Eq. \ref{eq:alpha-reward}). Increasing user satisfaction tends to decrease daily profit. Notably however in b), optimizing for user satisfaction has steered the agent to find policies that reduce the missing charge percentages while retaining a near-identical profit level. Data for a) is gathered over 20 training seeds with the shaded area representing standard deviation. In b-c) data is gathered per bar over 5 training seeds and 125 evaluations with the error bars again indicating standard devidation.}
    \label{fig:alpha_ablation}
\end{figure}

In Figure \ref{fig:learning_curve}, we have trained a standard PPO agent based on PureJaxRL \citep{lu2022discovered}. We trained on our included \textit{shopping} scenario in varying amounts of traffic using a 16 charger station (10 DC, 6 AC). We observe how our PPO agent increases its profit over a standard baseline. The baseline is set to always charge to its maximum potential within the constraints of the EVSE and the connected car. As expected, the potential for profit increases in scenarios with higher amounts of traffic, but this increase diminishes as we kept the charging station size the same.

Our baseline should yield a high customer satisfaction as customers should be charged within the minimum amount of possible time. 
In contrast, our charging station agent may optimize fully for short-term profit without consideration of user satisfaction. This is likely undesirable and may affect long-term profits. However, Chargax allows for flexible reward signals that may optimize for this. In Figure \ref{fig:alpha_ablation_a} and \ref{fig:alpha_ablation_b}, we trained our PPO agent to optimize for profit and user satisfaction at varying $\alpha$ levels. Notably in Figure \ref{fig:alpha_ablation_a}, we can see the agent manages to find preferential policies that substantially increase user satisfaction (decrease the amount of kWh that was not charged at departure time), while keeping profit levels quite similar.

\begin{table}[b] 
\caption{Performance comparison between Chargax and other EV charging Gym environments, based on data collected by performing 100k environment steps. We evaluated both taking random actions (assesing the performance of the transition function), and a training a PPO agent. The PPO agent was tested both with a single environment, and in a more typical training scenario with vectorized environments. Here we observe performance improvements of over 100x. The results are obtained on an \texttt{NVIDIA RTX 4000 Ada GPU} and an \texttt{AMD EPYC 2.8 GHz CPU}. For the comparison environments, we used Stable-Baselines3 \citep{raffin_stable-baselines3_2021} with CUDA enabled for the PPO implementation.} \label{tab:performance}
\begin{tabular}{l|c|cccccccc}
\toprule
 & Chargax & Ev2Gym & & Chargym & & Sustaingym & \\
 &  &  & \textbf{Speedup} &  & \textbf{Speedup} &  & \textbf{Speedup} \\
\midrule
Random & 1.36 & 77.95 & 57x & 36.34 & 27x & 1554.57 & 1144x \\
PPO (1) & 9.79 & 170.05 & 17x & 131.18 & 13x & 1718.71 & 176x \\
PPO (16) & 0.65 & 86.99 & \textbf{134x} & 125.06 & \textbf{192x} & 1836.00 & \textbf{2820x} \\
\bottomrule
\end{tabular}
\end{table}

\begin{figure}[t]
    \centering
    \includegraphics[width=\textwidth]{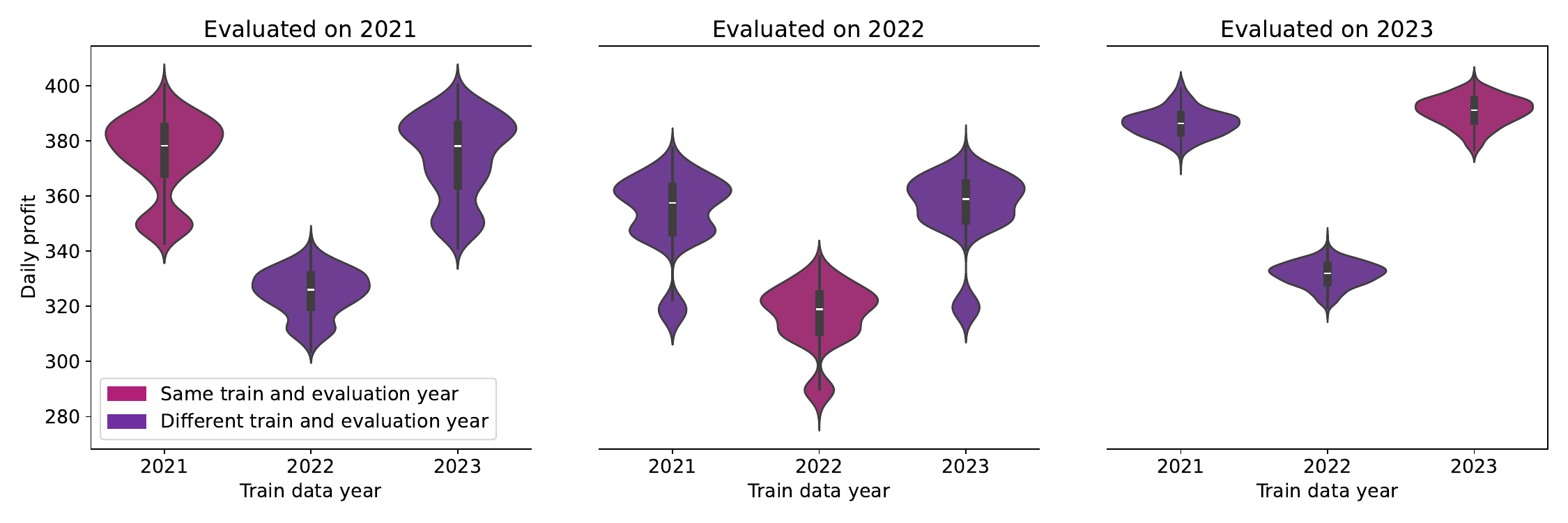}
    \caption{A PPO agent trained and tested on three separate years of Dutch electricity prices. For each of the three experiments, 10 agents with different seeds are trained on a fixed year (pink) and evaluated for 36500 episodes on all three years. Substantial price increases in the year 2022 results in suboptimal training when using this year's data -- even when evaluating on this same year.}
    \label{fig:dist-shift}
\end{figure}

Beyond finding appropriate reward signals, real-world deployment typically involves training an agent on historical exogenous data. During deployment, the agent likely encounters data that is has not yet observed. Possibly, the entire data set has shifted, for example, due to a rise in energy prices year-over-year. Therefore, it is important that system that deal with exogenous time-series data can deal with -- and test for -- this distribution shift \citep{yehSust}. As Chargax is flexibly designed to allow for any exogenous data, it readily allows to test for these distribution shift problems -- as is displayed in Figure \ref{fig:dist-shift}, where we have trained and evaluated RL agents on data of different price electricity years. Interestingly, although rewards would be assumed to peak when training and testing in the same year, employing data from 2021 or 2023 actually yielded higher rewards in 2022 compared to using the 2022 data directly. The EU region experienced significant energy price surges in 2022, likely complicating the training process with the data for this year.

Table \ref{tab:performance} and Figure \ref{fig:speedup}, showcases the performance of our environment compared to existing EV charging simulations that support reinforcement learning through a Gym API. We can see that in a typical training scenario, we can decrease learning times by factors exceeding 100.
It is important to acknowledge that these environments are not identical and might simulate different behaviours (for example, SustainGym does not allow discharging). Therefore, this comparison may be considered rough. However, the significant differences in scale clearly demonstrate the advantages of using Chargax- and JAX-based environments for RL in general. Training cycles can be reduced entire working days to well under 5 minutes, allowing for many more iterations of training and testing. 

\section{Discussion \& Conclusion}

This work presented Chargax, an EV charging simulator built in JAX. Chargax aims to bridge the gap between toy problems and real-world implementations, accelerating simulations while maintaining practical relevance. However, it remains a simulator, constrained by simplifying assumptions, requiring future work to further close the gap between simulation and deployment.

Our model assumes an isolated power network for the EV charging station, avoiding shared transformers that could introduce uncontrollable constraints. Expanding the model to include additional control variables, such as dynamic pricing strategies or vehicle allocation mechanisms, would increase its realism. Furthermore, accounting for temperature dependence in the system, or incorporating government-imposed regulatory constraints could make it more reflective of real-world charging stations. Furthermore, a natural addition for future work would be to incorporate local energy production systems (such as solar panels) and weather data.

In its current state, Chargax achieves training time reductions of over 100x, compared to existing simulators. Usual training durations of (multiple) working days can be completed in Chargax in well under 5 minutes, allowing for many more additional training and testing runs. We have built Chargax to be flexible, allowing for custom data sources for the exogenous state, and flexible reward structures. However, Chargax does provide base datasets and reward penalties to get started. 
While Chargax is inspired by a real-world scenario, it is designed as a general-purpose RL environment. Its speed, enabled by a fully JAX-based implementation, allows for efficient experimentation, and it supports the generation of varying sizes and complexities, enabling to test algorithms at various scales. Beyond EV charging, Chargax can be viewed as an instance of a broader class of optimal resource allocation problems.
We open source Chargax for the wider community to experiment with.$^{\ref{footnote:sourcecode}}$

\appendix

\section{Environmental Details} \label{ap:environment-details}

\subsection{State Spaces} \label{ap:state-space}

\paragraph*{Formal definition of exogenous and endogenous state space} To supplement the informal definitions of exogenous and endogenous state variables provided in the main text, we present formal definitions below.

Let $ s_t = (s_t^{(1)}, \dots, s_t^{(n)}) $ denote the state variables at time $ t $, and define the full history up to time $ t $ as $ h_t = (s_1, \dots, s_t) $. We further denote the history of the $ i $-th state variable as $ h_t^{(i)} = (s_1^{(i)}, \dots, s_t^{(i)}) $. Then, it is always possible to find a maximal index $ m \in \{0, \dots, n\} $ such that (possibly after reordering the state variables), the transition function can be factorized as follows:
\begin{equation}
    p(s_{t+1} \mid h_t, a_t) = p\left((s_t^{(1)}, \dots, s_t^{(m)}) \middle| (h_t^{(1)}, \dots, h_t^{(m)})\right) \cdot p\left((s_t^{(m+1)}, \dots, s_t^{(n)}) \middle| h_t, a_t\right).
\end{equation}
Based on this decomposition, we define the exogenous state space as $ s_{\mathrm{ex}} = (s^{(1)}, \dots, s^{(m)}) $ and the endogenous state space as $ s_{\mathrm{en}} = (s^{(m+1)}, \dots, s^{(n)}) $. Note that either component may be empty in the edge cases $ m = 0 $ or $ m = n $, respectively. This formalization ensures that the overall state space can always be partitioned into two distinct parts, with each state variable assigned to one of the two categories. 

\paragraph*{Exogenous state space}
\label{sec::AppendixCarModels}

Apart from price data, examples of exogenous state variables include power demand of the grid, weather data, or marginal operating emissions rate (MOER, \citealp{yehSust}), all of which could influence the maximisation objective but evolve according to some (agent independent) time series. It is important to note, that while the environment requires auxiliary data for most built-in reward functions, e.g. it is impossible to maximise profit without having access to prices, these exogenous state variables may be treated unobservable for the agent. On the contrary, one may add data to the exogenous state space, that is not required for any reward calculation, but may serve as additional learning signal, for instance day-ahead power prices.

Apart from the above examples arrival data, user profiles, and car profiles are part of the exogenous state space.

\begin{itemize}
    \item \textbf{Arrival Data} At each timestep $t$, a number of cars $M(t)$ is characterized as a sample from an arrival distribution $M(t) \sim \mathcal{D}_\mathrm{arrival}(t)$. 
    \item \textbf{Car Profiles}  Arriving cars are characterised by their physical properties. This encompasses the charging speed $\hat{r}$ as a function of the SoC. As in \citep{leeACNS2020} we assume a piece-wise linear function
    \begin{equation*}
        \hat{r}_{\tau, \Bar{r}}(\mathrm{SoC}) = 
        \begin{cases}
            \Bar{r}, & \mathrm{SoC} \leq \tau \\
            (1- \mathrm{SoC}) \frac{\Bar{r}}{1-\tau}, & \mathrm{SoC} > \tau.
        \end{cases}
    \end{equation*}
    Due to lack of data, we assume that the discharging speed can be obtained by vertically flipping the charging curve at $\mathrm{SoC}=0.5$.
    While we assume, that we have a different maximal charging speed for different charger types -- by default AC and DC charger -- and have hence different max charging rates ($\Bar{r} = (\Bar{r}_\mathrm{AC}, \Bar{r}_\mathrm{DC})$), we assume that both charging speed curves use the same $\tau$. Lastly, each car has a maximum battery capacity $C$, which is important for calculating State of Charges. These car profiles are sampled from a pre-defined car distribution $\mathcal{D}_\mathrm{car}(t)$, see also Table~\ref{table:data-overview}.
    \item \textbf{User Profiles} Additionally to the physical properties, the charging demand is a result from the habits of the car owner, encompassing a duration of stay $\Delta t_\mathrm{remain}$, the number of units of power to be charged $\Delta E$, the SoC upon arrival $\mathrm{SoC}_0$ and the user preference $u$, indicating whether a user is time-sensitive (will leave iff $\Delta t_\mathrm{remain}=0$), or charge sensitive (will leave iff $\Delta E_\mathrm{remain} = 0$). The user profiles are sampled from a distribution $\mathcal{D}_\mathrm{user}(t)$, see also Table \ref{table:data-overview}.
\end{itemize}

\paragraph*{Endogenous state space}  The endogenous state consists of the state of the various charging ports and their connected cars, and the station battery. For each charging port, we assume a fixed voltage and allow the actual power drawn to be regulated by controlling the current $I_{\mathrm{drawn}}(t) \in [0, I_\mathrm{max}]$ \citep{orfanoudakEV2G2024}. We assume that the voltage value already encodes the phases, i.e. it represents the product $V\cdot \sqrt{\phi}$ in \citet{orfanoudakEV2G2024}, eliminating the need for the phase as an additional variable. To incorporate losses during the (dis)charging process, each EVSE is equipped with an efficiency coefficient for charging and discharging. As a charging port may not always be occupied, we add a final Boolean to the state $\mathbbm{1}_\mathrm{occup}$, indicating the presence of a car. 

To properly facilitate discharging, the charging station is equipped with a battery. Similarly to EVSEs, the battery will have a fixed voltage $V_\mathrm{battery}$, with the power flow controlled by the current $I_\mathrm{battery}(t)$. To specify the physical properties of the battery, it also has a maximum capacity $C$, the maximal charging rate for a car $\Bar{r}$ and $\tau \in (0,1)$. Additionally, we will equip the state with the current SoC of the battery
\begin{equation*}
    s_\mathrm{battery} = (I_\mathrm{battery}(t), \mathrm{SoC}_\mathrm{battery}(t), \hat{r}_\mathrm{battery}(t)).
\end{equation*}

\paragraph*{Car State} Additionally, the state of each charging port contains information for the connected cars, the so-called car state, representing the car that is charging at this port (all zeros if no car is present). As this car state consists of exogenous and endogenous variables, it is listed separately. This includes the car's state-of-charge (SoC $\in [0,1]$),  the remaining required power $\Delta E_\mathrm{remain} \in \R_{\geq 0}$, the number of timesteps the car remains  $\Delta t_\mathrm{remain}\in \N$, and the maximal charging power currently allowed by the car $\hat{r}(t)\in \R_{\geq 0}$. The latter one is heavily depended on the State of Charge $\mathrm{SoC}(t) \in [0,1]$ of the car battery \citep{welzelGrid2021, fastned}, which is also part of the car-state.
The car-state also contains information about the physical properties of the car. These are the maximum battery capacity $C$, the maximum charging rate for a car $\Bar{r}$, and $\tau \in (0,1)$ -- the transition point from the bulk stage to the absorption stage of the charging process \citep{leeACNS2020}. Finally, the car-state includes a user preference indicator $u$.

In brief summary, the state of each charging port is represented by:
\begin{itemize}
    \item Current power drawn $I_{\mathrm{drawn}}(t)\in \R_{\geq 0}$, occupancy indicator $\mathbbm{1}_\mathrm{occup}(t) \in \{0,1\}$;
    \item Car-state $(\Delta E_\mathrm{remain}(t), \Delta t_\mathrm{remain}(t), 
    \hat{r}(t), \mathrm{SoC}(t), C, \Bar{r}, \tau, u)$.
\end{itemize}

\subsection{Transition Function} \label{ap:transition-function}

The transition function consists of four major steps: (i) Apply Actions, i.e. adapt charging levels at each EVSE, (ii) charge stationed cars, (iii) departure of cars, and (iv) arrival of new cars.

\paragraph*{Apply Actions} As a first step, the actions taken by the agent are applied to adjust the power drawn by each charging pole, specifically
\begin{equation*}
        I_{\mathrm{drawn}, i}(t) = 
        \begin{cases}
            \min\left(I_{\mathrm{drawn}, i}(t - \Delta t) + a_i(t) ,\hat{r}(t), I_\mathrm{max\to, i}\right) & I_{\mathrm{drawn}, i}(t - \Delta t) + a_i(t) \geq 0\\
            -\min\left(-I_{\mathrm{drawn}, i}(t - \Delta t) - a_i(t) ,\hat{r}(t), I_\mathrm{max\leftarrow, i}\right) & \mathrm{else}.
        \end{cases}
    \end{equation*}
Hereby constraints on the maximum power drawn imposed by the architecture are enforced by assuring that for each subtree $H$ in the architecture, the constraints
\begin{equation}
    \label{eq:ArchConst}
    \frac{1}{\eta_H} \sum_{h \in \mathrm{leaves}(H)} I_{\mathrm{drawn}, h}(t) \leq I_H,
\end{equation}
are satisfied. If the drawn currents violate these constraints, the currents at each leaf are rescaled to satisfy the constraints, modelling the potential behaviour of some safety infrastructure on top of the controller. 

\paragraph*{Charge Stationed Cars} After having adjusted the power levels at each charging pole, the charging is processed for the time interval, where a constant charging rate over the full interval $\Delta t$ is assumed. The car states are adjusted in the following way:
\begin{align*}
    \Delta E_{\mathrm{remain},i}(t+\Delta t) &= \Delta E_{\mathrm{remain},i}(t) - \Delta t \cdot V_i \cdot I_{\mathrm{drawn}, i}(t) \\
    \mathrm{SoC}(t+\Delta t) &= \mathrm{SoC}(t) + \frac{\Delta t \cdot V_i \cdot I_{\mathrm{drawn}, i}(t)}{C_i} \\
    \hat{r}(t+\Delta t) &= \hat{r}_{\tau_i, \Bar{r}_{i}}(\mathrm{SoC}(t+\Delta t)) .
\end{align*}
Notably, the physical attributes of the car in the car state, i.e. the maximum battery capacity, the maximal charging rate and $\tau$ do not change. As charging has been proceed, we assume that time moves on, i.e. $t \mapsto t + \Delta t$ and  $\Delta t_{\mathrm{remain}, i}(t+\Delta t) = \Delta t_{\mathrm{remain}, i}(t) - \Delta t$.

\paragraph*{Departure of Cars} At the end of the period, cars fully charged or with no time remaining will leave. Consequently the car-states for the corresponding charging poles are updated 
\begin{equation*}
    s_{c,i}(t) = 
    \begin{cases}
        (0,\dots,0) & \Delta t_{\mathrm{remain},i}(t) = 0 \,\text{and}\, u_i = 0\\
        (0,\dots,0) & \Delta E_{\mathrm{remain},i}(t) = 0 \,\text{and}\, u_i = 1\\
        s_{c,i}(t) & \text{else}. 
    \end{cases}
\end{equation*}

\paragraph*{Arrival of new Cars} The amount of arriving cars is sampled $M(t)\sim \mathcal{D}_\mathrm{arrival}(t)$. We model a first-come-first-served policy by clipping $M(t)$ by the number of available free spots $N - \sum_{i=1}^N \mathbbm{1}_{\mathrm{occup}, i}(t)$. For each car $j=1,\dots,M(t)$ the car profile, and the user profile are sampled from their respective distribution, i.e. $(\Delta t_{\mathrm{remain}, j}, \,\Delta E_j, \,\mathrm{SoC}_{0, j}, u_j) \sim \mathcal{D}_\mathrm{profile}(t)$ and $(\Bar{r}_{j}, \tau_j, C_j) \sim \mathcal{D}_\mathrm{car}(t)$, respectively. 

Each car $j$ is then allocated to a free charging pole $k$, which alters the state of charging pole $k$ based on car $j$:
\begin{equation*}
    s_{c,k}(t) = (0, 1, \Delta E_j, \Delta t_{\mathrm{remain}, j}, \hat{r}_{\tau_j, \Bar{r}_{j}}(\mathrm{SoC}_{0,j}), C_j, \Bar{r}_{j}, \tau_j, u_j).
\end{equation*}

\subsection{Reward functions}
\label{ap:rewards}

The amount of net energy transferred into cars in the interval $[t,t+\Delta t]$ can be calculated as $\Delta E_\mathrm{net}(t) = \Delta t \sum_{i=1}^N V_i \cdot I_{\mathrm{drawn}, i}(t)$. Accounting for losses within the electric architecture of the charging station, the amount of energy, that is transferred from cars into the grid can be calculated as 
\begin{equation}
    \label{eq::Profit1}
    \Delta E_{\to \mathrm{grid}}(t) = \Delta t \sum_{i: I_{\mathrm{drawn}, i} < 0} \eta_i \cdot V_i \cdot I_{\mathrm{drawn}, i}(t) < 0.
\end{equation}
Similarly, the amount of energy that has to be drawn from the net to transfer set levels of energy into cars $\Delta E_{\mathrm{grid} \to}(t)$, after incorporating imperfect efficiencies, can be calculated via $\Delta E_{\mathrm{grid} \to}(t) = \Delta t \sum_{i: I_{\mathrm{drawn}, i} > 0} \eta_i^{-1}\cdot V_i \cdot I_{\mathrm{drawn}, i}(t) >0$. Lastly, the energy contributed by (dis-)charging the battery $\Delta E_\mathrm{b, net}(t) = \Delta t\,I_\mathrm{battery}(t) \,V_\mathrm{battery}$ has to be incorporated, resulting in the following net energy drawn from (or pushed into) the grid
\begin{equation*}
    \Delta E_{\mathrm{grid, net}} = \Delta E_{\mathrm{grid} \to}(t) + \Delta E_{\to \mathrm{grid}}(t) + \Delta E_\mathrm{b, net}(t).
\end{equation*}
Further that the price at which we sell and buy power from car owners is the same, i.e. $p_\mathrm{sell}$. This results in the following revenue
\begin{equation*}
    \Pi(t) = \begin{cases}
        p_\mathrm{sell}(t) \cdot \Delta E_\mathrm{net}(t) - p_\mathrm{buy}(t) \cdot \Delta E_{\mathrm{grid, net}} - c_{\Delta t}& \Delta E_{\mathrm{grid, net}} > 0, \\
        p_\mathrm{sell}(t) \cdot \Delta E_\mathrm{net}(t) - p_\mathrm{sell, grid}(t) \cdot \Delta E_{\mathrm{grid, net}} - c_{\Delta t} & \Delta E_{\mathrm{grid, net}} \leq 0.
    \end{cases}
\end{equation*}
Here, $c_{\Delta t}$ denotes the fixed cost for running the facility per $\Delta t$. The general reward $r(s(t),a(t),s(t+\Delta t))$, abbreviated by $r(t)$ in Chargax consists of the profit minus the linear combination of some penalty terms
\begin{equation}
    r(t) = \Pi(t) - \sum_c \alpha_c c(t).
\end{equation}
Some examples of included penalty terms are listed below
\begin{itemize}
    \item \textbf{Constraint Violations} The hard constraints imposed by the architecture in Eq. \ref{eq:ArchConst} could be instead included as as soft constraints \citep{yehSust} via the penalty
    \begin{equation*}
        c_\mathrm{constraint}(t) = \max_H\min\left(0, \frac{1}{\eta_H} \sum_{i \in \mathrm{leaves}(H)} I_{\mathrm{drawn}, i}(t) - I_H\right).
    \end{equation*}
    \item \textbf{Satisfaction penalty} Users can experience dissatisfaction in two ways: Time-sensitive users have a desired departure time and are assumed to leave at that time, regardless the SoC of their car. To avoid customers leaving the charging station with a suboptimal SoC we propose to incorporate a satisfaction penalty
    \begin{equation*}
        c_\mathrm{Satisfcation, 0}(t) = \sum_{i: \Delta t_{\mathrm{remain}, i}(t) = 0, u_i=0} \max(0, \Delta E_\mathrm{remain, i}(t)).
    \end{equation*}
    The opposite holds for charge sensitive users, as they are expected to leave when there cars are charged to the desired level. However, these users can be overly satisfied by charging their car to the desired level faster than desired
    \begin{equation*}
        c_\mathrm{Satisfcation, 1}(t) = \sum_{i: \Delta E_{i}(t) = 0, u_i=1} \max(0, - \Delta t_{\mathrm{remain}, i}(t)) - \beta \max(0, \Delta t_{\mathrm{remain}, i}(t)).
    \end{equation*}
    Here $\beta$ controls how much the positive satisfaction from leaving earlier should weight in comparison to the negative dissatisfaction from having to stay overtime.    
    \item \textbf{Sustainability} To enforce the agent to charge cars in the most sustainable way possible, a penalty term for non-sustainable behaviour may be added. One solution proposed in \citep{yehSust} is to employ the MOER $m(t)$, capturing the carbon intensity of a unit of energy produced at time $t$
    \begin{equation*}
        c_\mathrm{sustain}(t) = m(t) \cdot \Delta E_\mathrm{grid, net}(t).
    \end{equation*}
    \item \textbf{Rejected Customers} In view of congestion management problems \citep{zhangOpti2019, hussainOpti2022}, one might be interested in serving the maximum number of cars, i.e. reduce the amount of rejected cars, by adding a penalty term for declined cars
    \begin{equation*}
        c_\mathrm{declined}(t) = \max\left(M(t) - \left(N - \sum_{i=1}^N \mathbbm{1}_{\mathrm{occup}, i}(t)\right), 0\right).
    \end{equation*}
    \item \textbf{Battery Degradation} Real world batteries suffer from degradation under use \citep{leeAnal2020}. This can be incorporated by adding a degradation cost to every discharging of the Charging station battery, as well as for the cars. For sake of simplicity, we assume the additional degradation to be proportional to the discharged energy
    \begin{equation*}
        c_\mathrm{degrad, battery}(t) = |\Delta E_\mathrm{b, net}(t)| \cdot\mathbbm{1}_{\{\Delta E_\mathrm{b, net}(t) < 0\}} \,\,\text{and}\,\,c_\mathrm{degrad, cars}(t) = |\Delta E_\mathrm{\rightarrow grid}(t)| .
    \end{equation*}
    \item \textbf{Grid Stability} (Only applicable in a V2G scenario) If the agent can discharge cars, this can be leveraged to stabilize the grid load \citep{liCoor2021a, elmaDyna2020}.
    This could be reflected in a penalty term through an exogenous signal of the grid demand $d_\mathrm{grid}(t) \in \R$
    \begin{equation*}
        c_\mathrm{grid}(t) = | \Delta E_\mathrm{net}(t) - d_\mathrm{grid}(t)|.
    \end{equation*}
\end{itemize}

\subsubsection*{Acknowledgments}
\label{sec:ack}
The authors thank Joost Commandeur for his invaluable feedback during the process of this work. This work was supported by Shell Information Technology International Limited and the Netherlands Enterprise Agency under the grant PPS23-3-03529461.


\bibliography{main}
\bibliographystyle{rlj}

\beginSupplementaryMaterials

\section{Implementation Details}
\label{sec:AppendixImplementationDetails}

\subsection{Practical Considerations}

Table \ref{tab:ppo-config} contains environment settings used throughout our experiments whenever not stated. Additionally, we list some practical considerations in Chargax here. 
\begin{itemize}
    \item The episode length defaults to the length of data provided for arriving cars. In our bundled scenarios, this equals 24 hours. These bundled scenarios provide their data as average numbers per timestep. The actual number of cars arriving is then drawn using a Poisson distribution.
    \item By default, we train in a Chargax environment utilizing a method akin to exploring starts. At environment reset, we sample a random day from the given price data and use this day's prices for the episode. The agent observes the current episode day and whether this is a weekday or a workday. 
    \item Throughout our experiments, we have used a discretised action space, setting the (user-defined) discretization level to 10. This allows the agent to select increments as 10\%, 20\%, 30\%, etc., up to 100\% of the maximum current for each charging port.

\end{itemize}

\subsection{Agent configuration}
Unless otherwise stated, the experiments conducted in Section \ref{sec:experiments} and Appendix \ref{sec:AppendixAdditionalExperiments} trained with a PPO agent using the hyperparameters listed in Table \ref{tab:ppo-config}.

\begin{table}[h]
    \centering
    \begin{tabular}{ll||ll}
        \toprule
        Hyperparameter & Value & Environment Parameter & Value \\ 
        \midrule
        Total timesteps & 1e7 & Minutes per timestep  $\Delta t$ & 5 \\ 
        Learning rate ($\alpha$) & 2.5e-4 (annealed) & Discretization factor & 10 \\ 
        Discount factor $\gamma$ & 0.99 & Episode length & 24 hours \\ 
        GAE $\lambda$ & 0.95 & Number of Chargers & 16 \\ 
        Max grad norm & 100.0 & Number of DC Chargers & 10 \\
        Clipping coefficient $\epsilon$ & 0.2 & Sell price to customers ($p_\mathrm{sell}$) & 0.75 \\
        Value func clip coefficient & 10.0 & All reward coefficients $\alpha$ (Eq. \ref{eq:alpha-reward}) & 0.0 \\
        Entropy coefficient & 0.01 & & \\
        Value function coefficient & 0.25 & & \\
        Vectorized environments & 12 & & \\
        Rollout length (steps) & 300 & & \\
        Number of minibatches & 4 & & \\
        Update epochs & 4 & & \\
        Minibatch size & 900 & & \\
        Batch size & 3600 & & \\
        \bottomrule
    \end{tabular}
    \caption{PPO hyperparameters (left) alongside environment settings (right) used throughout our experiments unless otherwise stated.}
    \label{tab:ppo-config}
\end{table}

\newpage

\section{State summary}

\begin{table}[htbp]
\caption{Summary of the state space in Chargax}
\begin{tabular}{c|c|c|c|c}
\multicolumn{1}{l|}{}                                                                & symbol                          & domain        & \begin{tabular}[c]{@{}c@{}}exogenous/\\ endogeneous\end{tabular} & variable name                        \\ \hline
\multicolumn{1}{l|}{\multirow{5}{*}{reward data}}                                    & $p_\mathrm{sell}$               & $\R_{\geq 0}$ & exogenous                                                        & Selling price (to Customer) per kWh  \\
\multicolumn{1}{l|}{}                                                                & $p_\mathrm{buy}$                & $\R_{\geq 0}$ & exogenous                                                        & Buying price per kWh                 \\
\multicolumn{1}{l|}{}                                                                & $p_\mathrm{sell, grid}$         & $\R_{\geq 0}$ & exogenous                                                        & Selling price (to grid) per kWh      \\
\multicolumn{1}{l|}{}                                                                & $m$                             & $\R_{\geq 0}$ & exogenous                                                        & Marginal Operations Emission Rate    \\
\multicolumn{1}{l|}{}                                                                & $d_\mathrm{grid}$               & $\R$          & exogenous                                                        & Grid Demand                          \\ \hline
\multicolumn{1}{l|}{}                                                                & $M$                             & $\N_{0}$      & exogenous                                                        & Number of arriving cars              \\ \hline
\multirow{8}{*}{\begin{tabular}[c]{@{}c@{}}Car state of \\ EVSE i\end{tabular}}      & $\Delta t_\mathrm{remain,i}$    & $\N_{0}$      & exogenous                                                        & Remaining time of customer           \\
                                                                                     & $C_i$                           & $\R_{\geq 0}$ & exogenous                                                        & Capacity of Car                      \\
                                                                                     & $\bar{r}_i$                     & $\R_{\geq 0}$ & exogenous                                                        & Maximum charging rate                \\
                                                                                     & $\hat{r}_i$                     & $\R_{\geq 0}$ & exogenous                                                        & Maximum charging rate at current SoC \\
                                                                                     & $\tau_i$                        & $[0,1]$       & exogenous                                                        &                                      \\
                                                                                     & $u_i$                           & $\{0,1\}$     & exogenous                                                        & User preference                      \\
                                                                                     & $\mathrm{SoC}_i$                & $[0,1]$       & endogenous                                                       & Current SoC                          \\
                                                                                     & $\Delta E_\mathrm{remain,i}$    & $\R_{\geq 0}$ & endogenous                                                       & Remaining Charging demand            \\ \hline
\multirow{2}{*}{\begin{tabular}[c]{@{}c@{}}State variables\\ of EVSE i\end{tabular}} & $\mathbbm{1}_\mathrm{occup,i}$  & $\{0,1\}$     & endogenous                                                       & Occupancy Indicator                  \\
                                                                                     & $I_\mathrm{drawn, i}$           & $\R_{\geq 0}$ & endogenous                                                       & Current Power drawn at EVSE          \\ \hline
\multirow{3}{*}{Battery state}                                                       & $I_\mathrm{battery}$            & $\R_{\geq 0}$ & endogenous                                                       & Current power drawn at battery       \\
                                                                                     & $\mathrm{SoC}_\mathrm{battery}$ & $[0,1]$       & endogenous                                                       & SoC of Battery                       \\
                                                                                     & $\hat{r}_\mathrm{battery}$      & $\R_{\geq 0}$ & endogenous                                                       & Maximum charging rate at current SoC
\end{tabular}
\end{table}

\section{Additional Experiments} 
\label{sec:AppendixAdditionalExperiments}

\begin{figure}[ht]
    \centering
    \begin{subfigure}[b]{0.245\textwidth} 
        \centering
        \includegraphics[width=\textwidth]{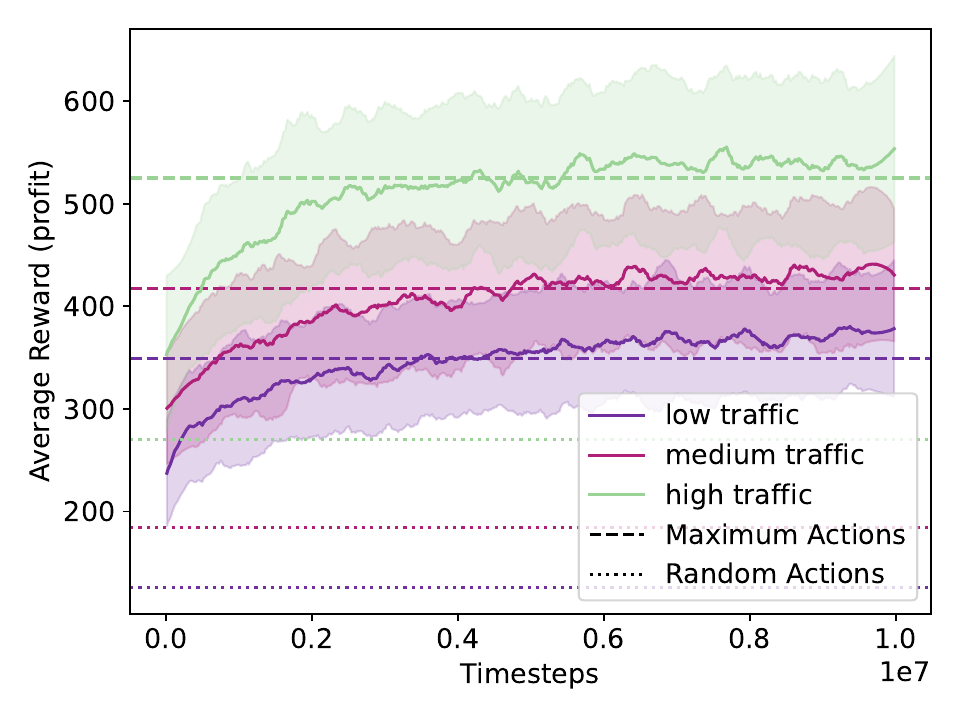}
        \caption{Residential}
    \end{subfigure}
    \hfill
    \begin{subfigure}[b]{0.245\textwidth} 
        \centering
        \includegraphics[width=\textwidth]{images/shopping_high.pdf}
        \caption{Shopping}
    \end{subfigure}
    \hfill
    \begin{subfigure}[b]{0.245\textwidth} 
        \centering
        \includegraphics[width=\textwidth]{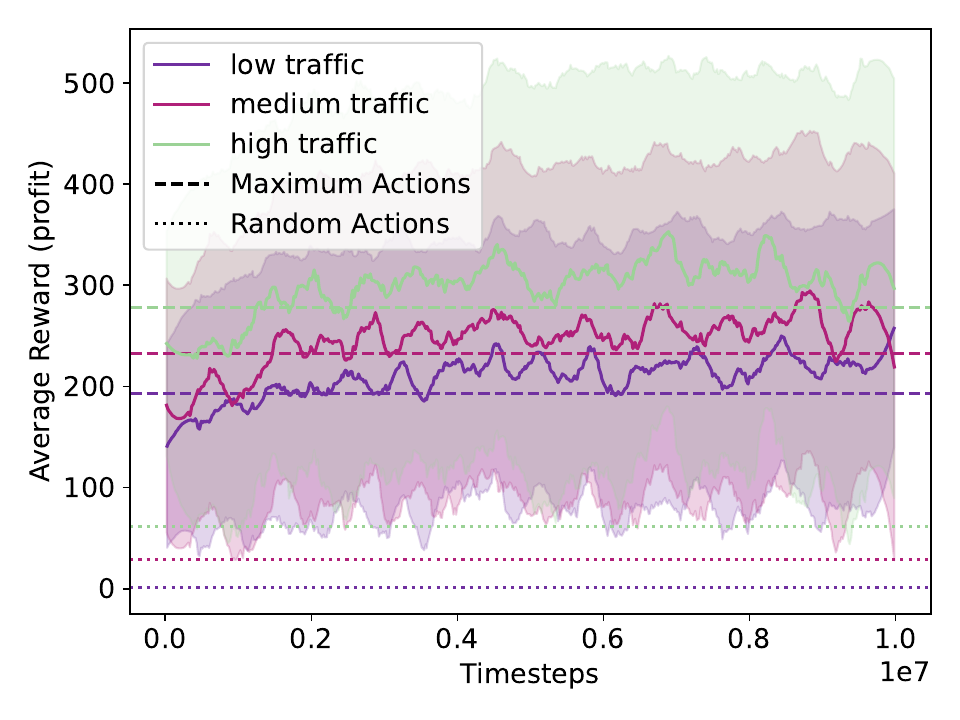}
        \caption{Workplace}
    \end{subfigure}
    \hfill
    \begin{subfigure}[b]{0.245\textwidth} 
        \centering
        \includegraphics[width=\textwidth]{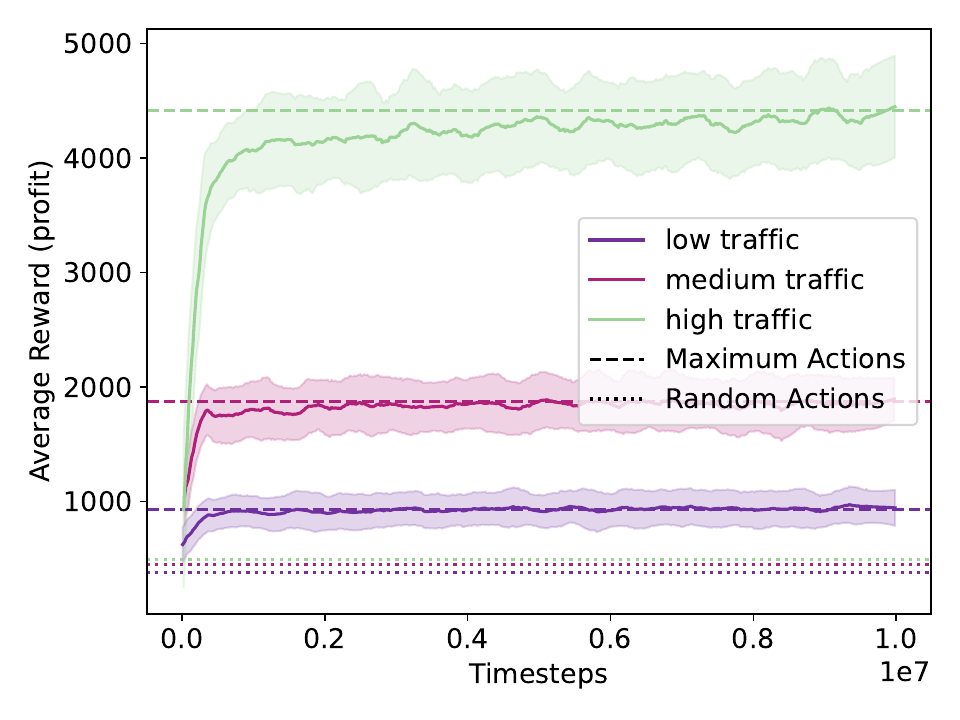}
        \caption{Highway}
    \end{subfigure}
    \hfill
    \caption{Results on our 4 bundled scenarios using \textbf{EU} cars and 16 chargers (10 DC, 5 AC)}
\end{figure}

\begin{figure}[ht]
  \centering
  \begin{subfigure}{0.245\textwidth}
    \centering
    \includegraphics[width=\textwidth]{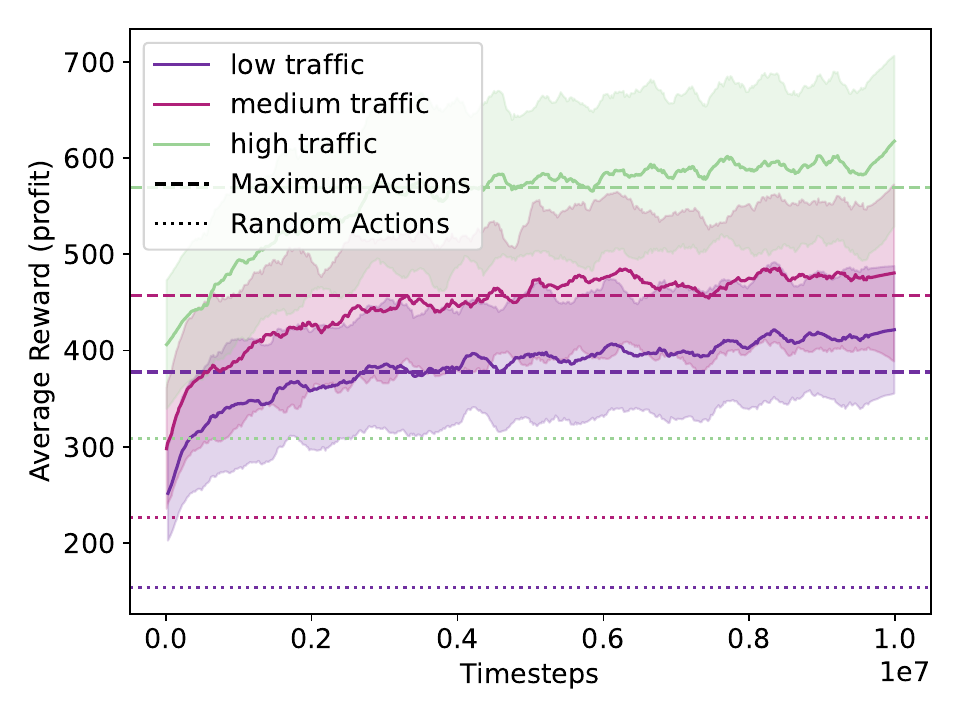}
    \caption{Residential}
  \end{subfigure}
  \hfill
  \begin{subfigure}{0.245\textwidth}
    \includegraphics[width=\textwidth]{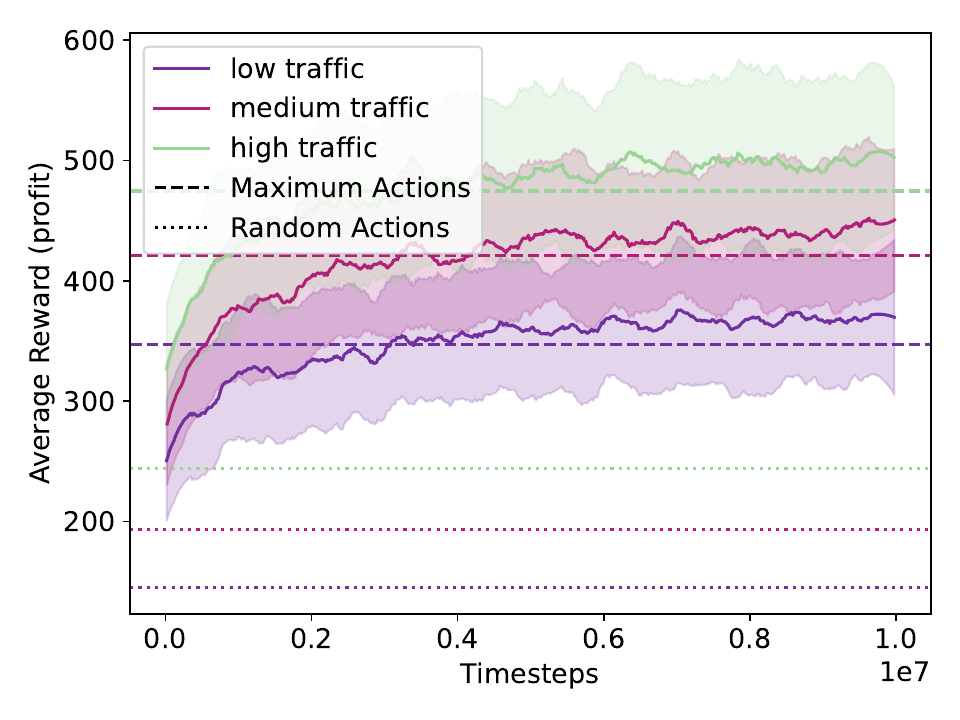}
    \caption{Shopping}
  \end{subfigure}
  \hfill
  \begin{subfigure}{0.245\textwidth}
    \centering
    \includegraphics[width=\textwidth]{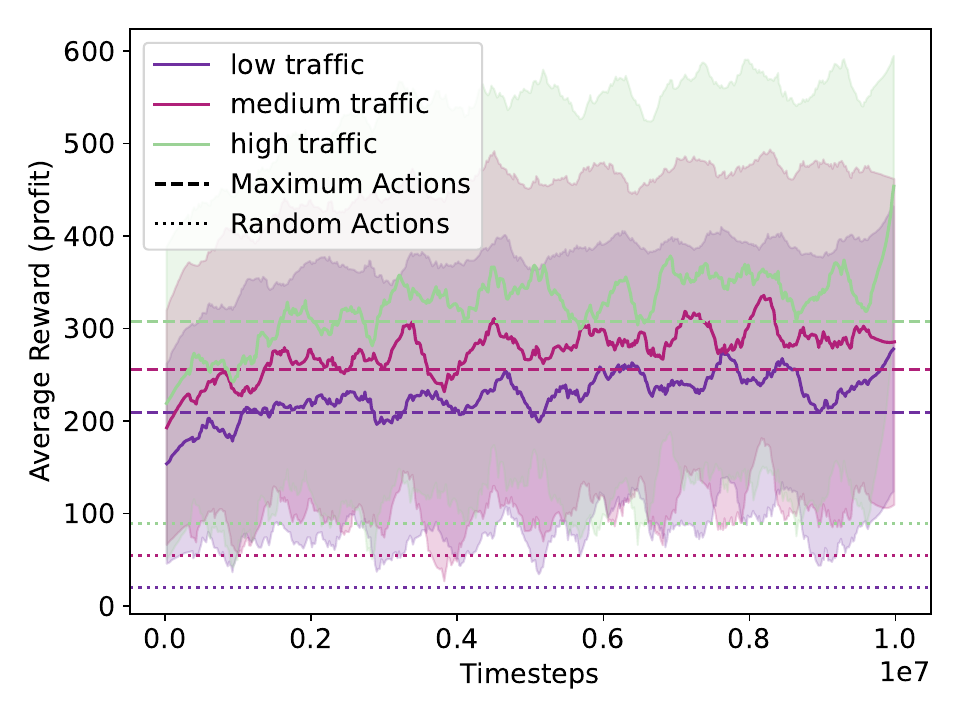}
    \caption{Workplace}
  \end{subfigure}
  \hfill
  \begin{subfigure}{0.245\textwidth}
    \centering
    \includegraphics[width=\textwidth]{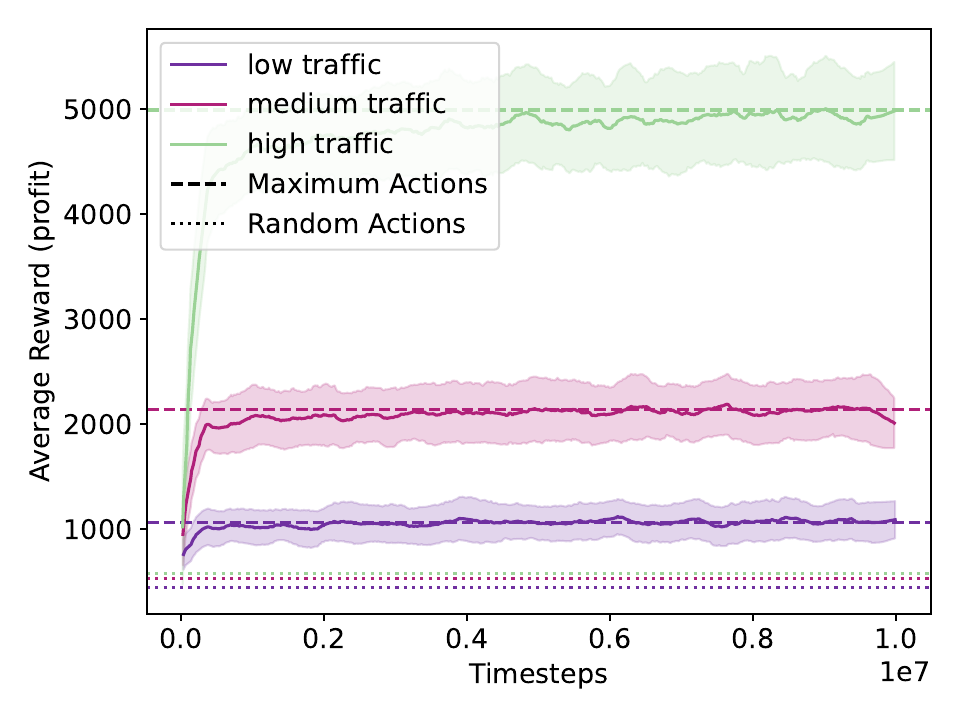}
    \caption{Highway}
  \end{subfigure}
  \caption{Results on our 4 bundled scenarios using \textbf{US} cars and 16 chargers (10 DC, 5 AC)}
\end{figure}

\begin{figure}[ht]
  \centering
  \begin{subfigure}{0.245\textwidth}
    \centering
    \includegraphics[width=\textwidth]{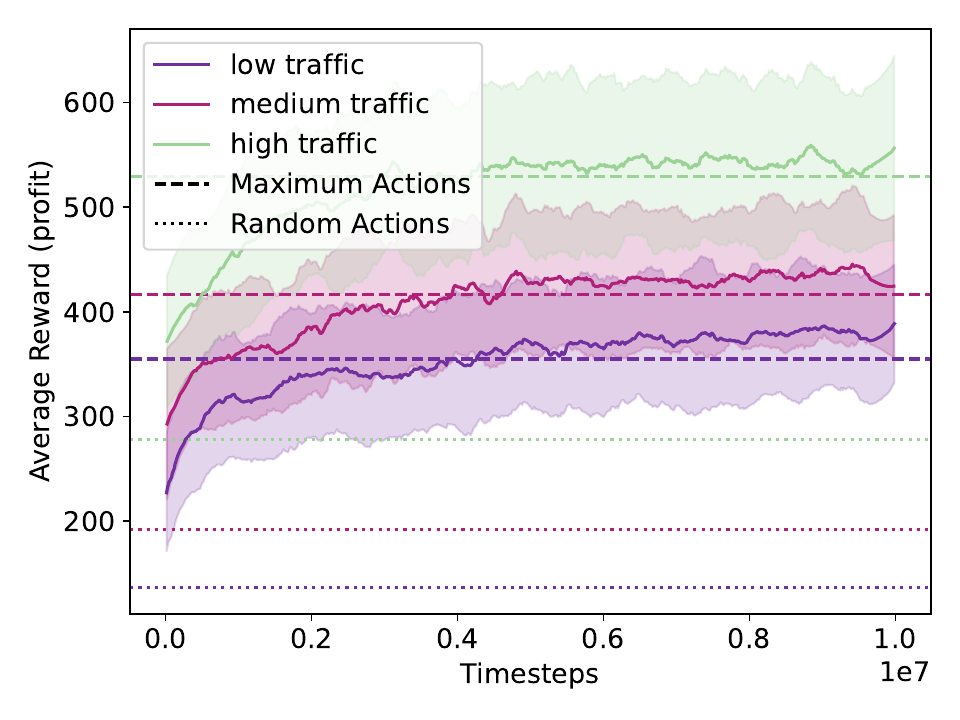}
    \caption{Residential}
  \end{subfigure}
  \hfill
  \begin{subfigure}{0.245\textwidth}
    \centering
    \includegraphics[width=\textwidth]{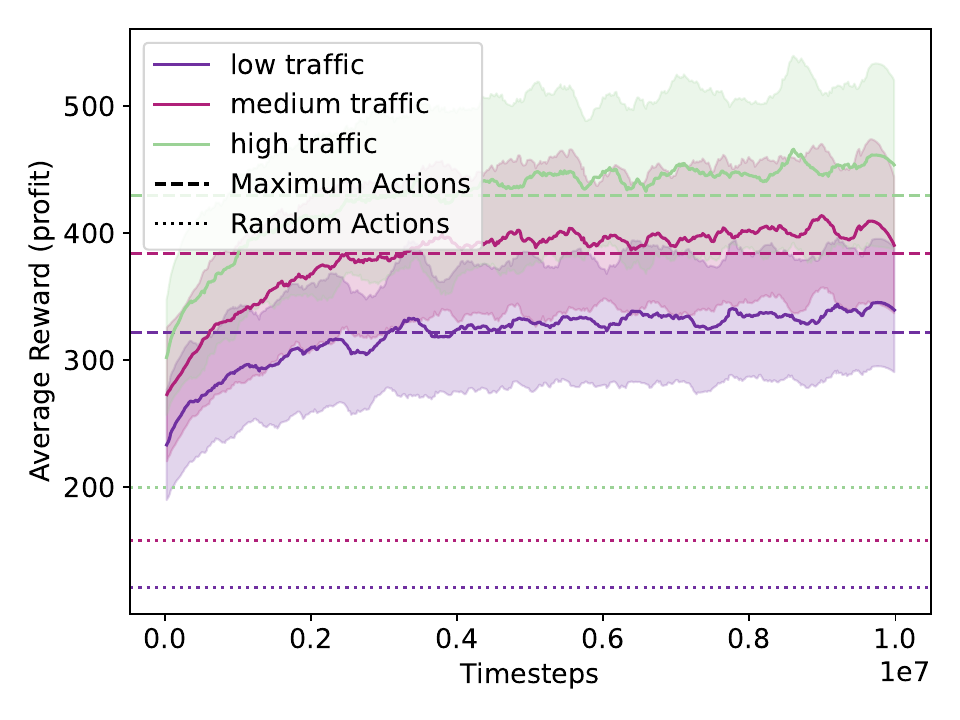}
    \caption{Shopping}
  \end{subfigure}
  \hfill
  \begin{subfigure}{0.245\textwidth}
    \centering
    \includegraphics[width=\textwidth]{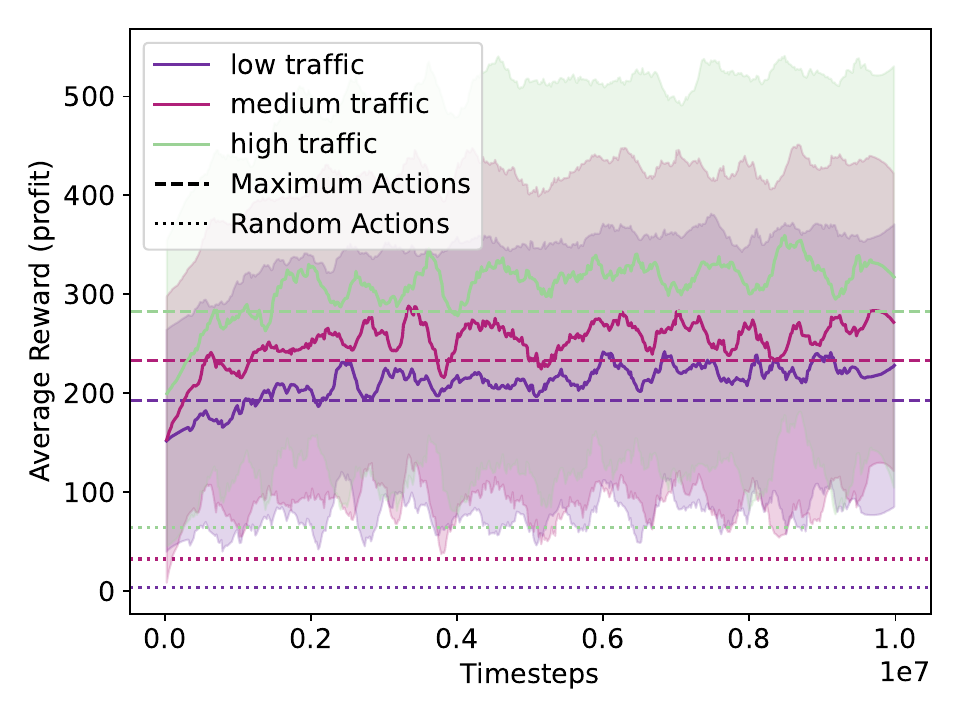}
    \caption{Workplace}
  \end{subfigure}
  \hfill
  \begin{subfigure}{0.245\textwidth}
    \centering
    \includegraphics[width=\textwidth]{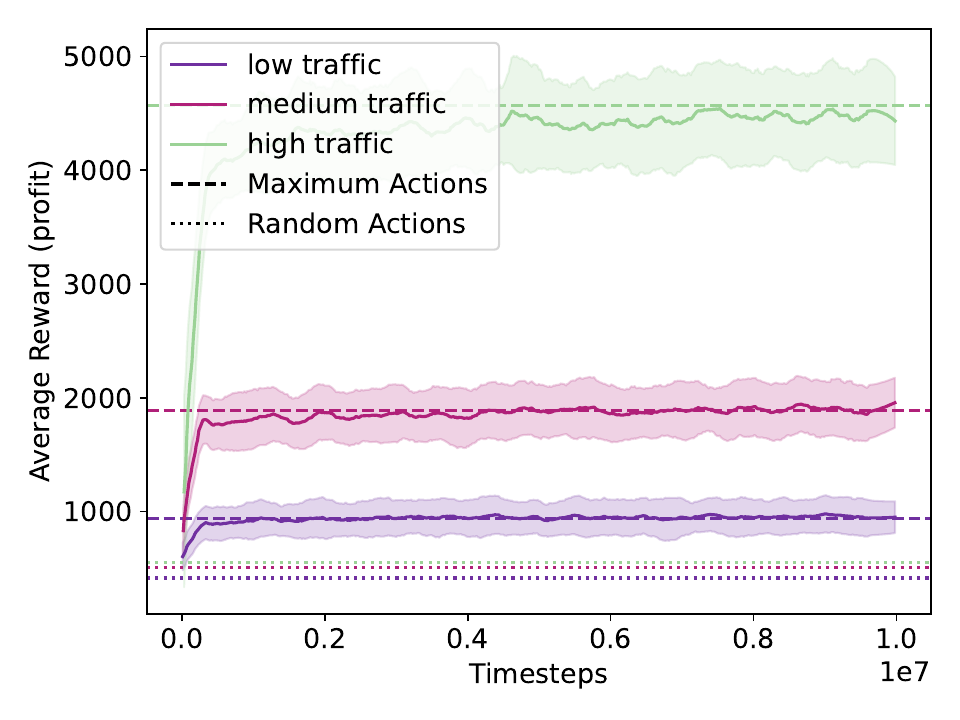}
    \caption{Highway}
  \end{subfigure}
  \caption{Results on our 4 bundled scenarios using \textbf{World} cars and 16 chargers (10 DC, 5 AC)}
\end{figure}


\begin{figure}[ht]
  \centering
  \begin{subfigure}{0.245\textwidth}
    \centering
    \includegraphics[width=\textwidth]{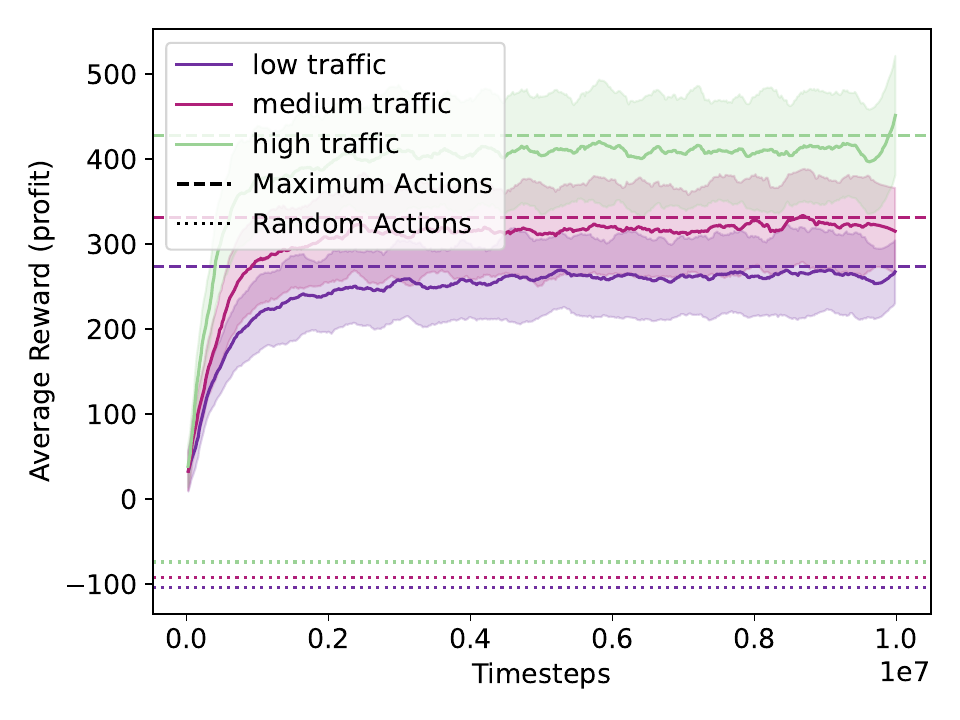}
    \caption{Residential}
  \end{subfigure}
  \hfill
  \begin{subfigure}{0.245\textwidth}
    \centering
    \includegraphics[width=\textwidth]{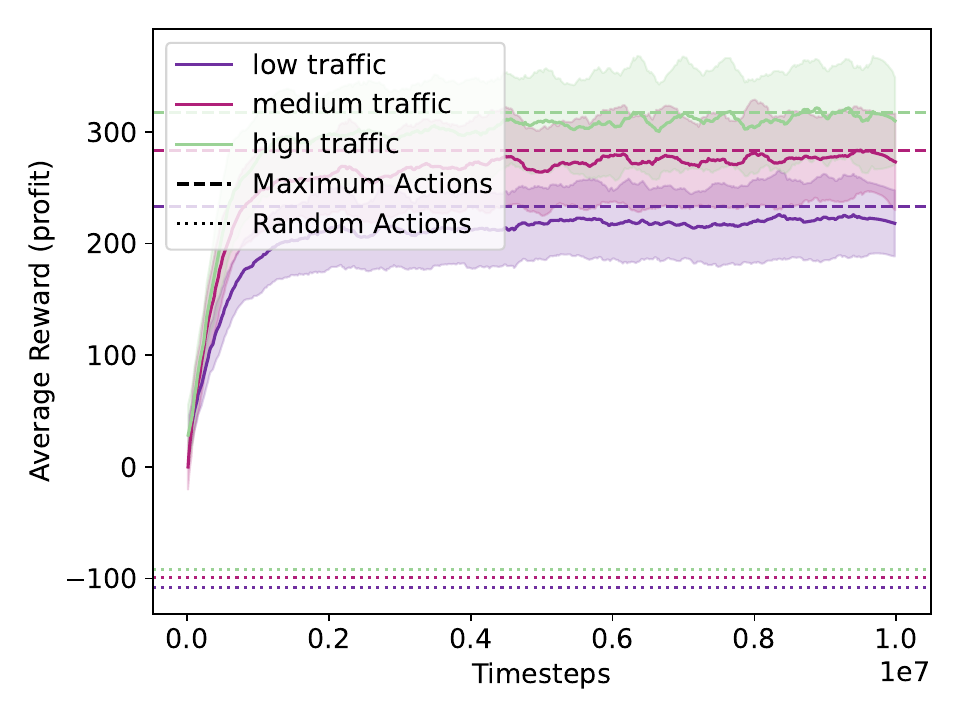}
    \caption{Shopping}
  \end{subfigure}
  \hfill
  \begin{subfigure}{0.245\textwidth}
    \centering
    \includegraphics[width=\textwidth]{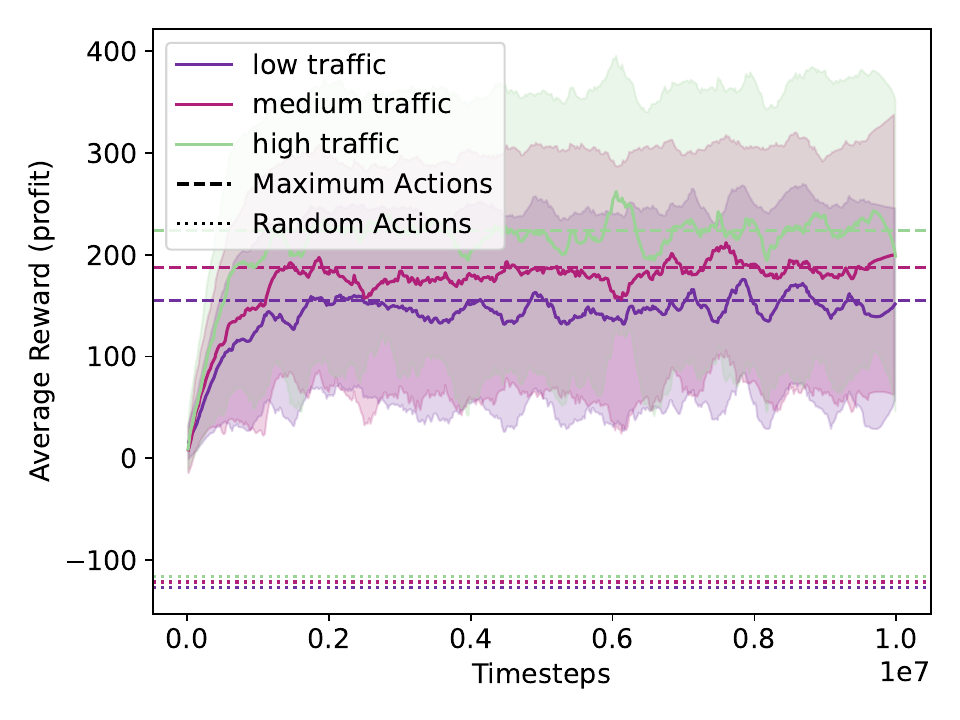}
    \caption{Workplace}
  \end{subfigure}
  \hfill
  \begin{subfigure}{0.245\textwidth}
    \centering
    \includegraphics[width=\textwidth]{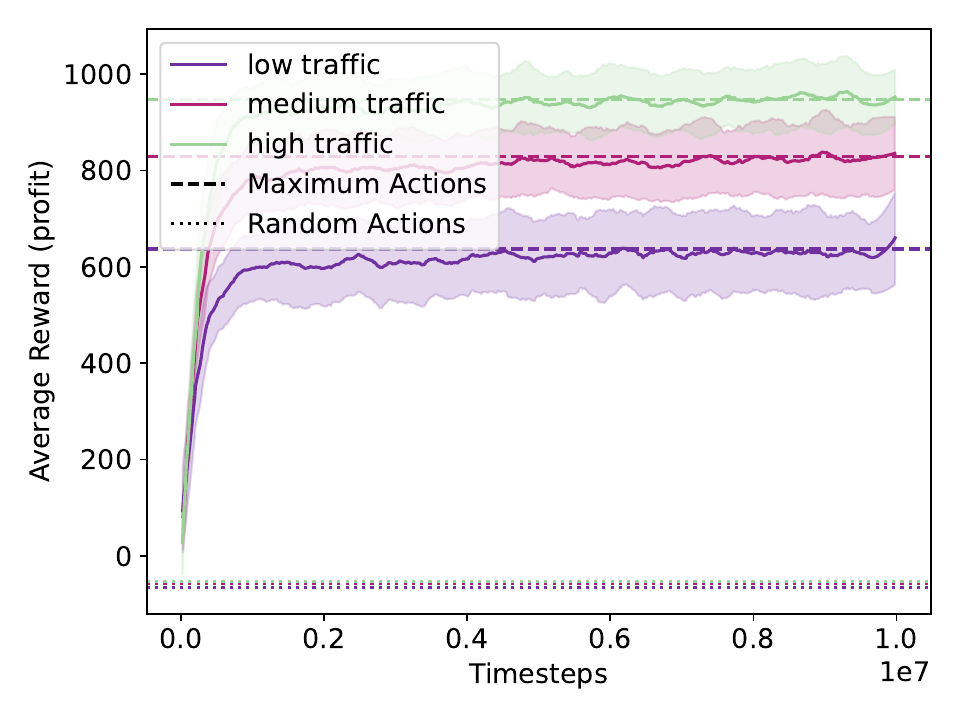}
    \caption{Highway}
  \end{subfigure}
  \caption{Results on our 4 bundled scenarios using EU cars and 16 AC (11.5kW) chargers}
\end{figure}

\begin{figure}[ht]
  \centering
  \begin{subfigure}{0.245\textwidth}
    \centering
    \includegraphics[width=\textwidth]{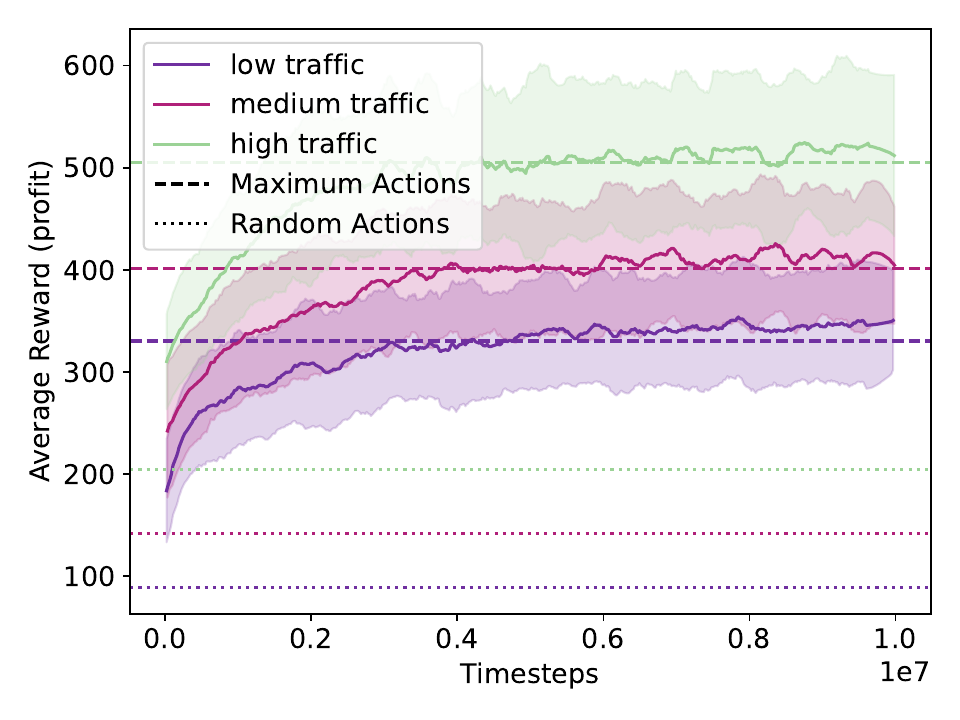}
    \caption{Residential}
  \end{subfigure}
  \hfill
  \begin{subfigure}{0.245\textwidth}
    \centering
    \includegraphics[width=\textwidth]{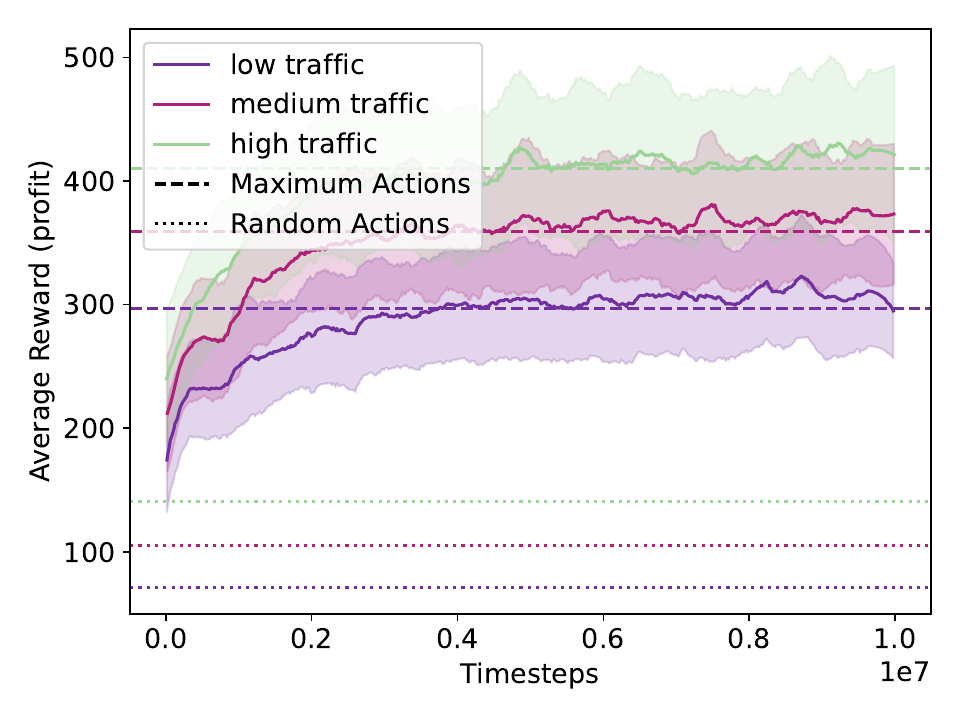}
    \caption{Shopping}
  \end{subfigure}
  \hfill
  \begin{subfigure}{0.245\textwidth}
    \centering
    \includegraphics[width=\textwidth]{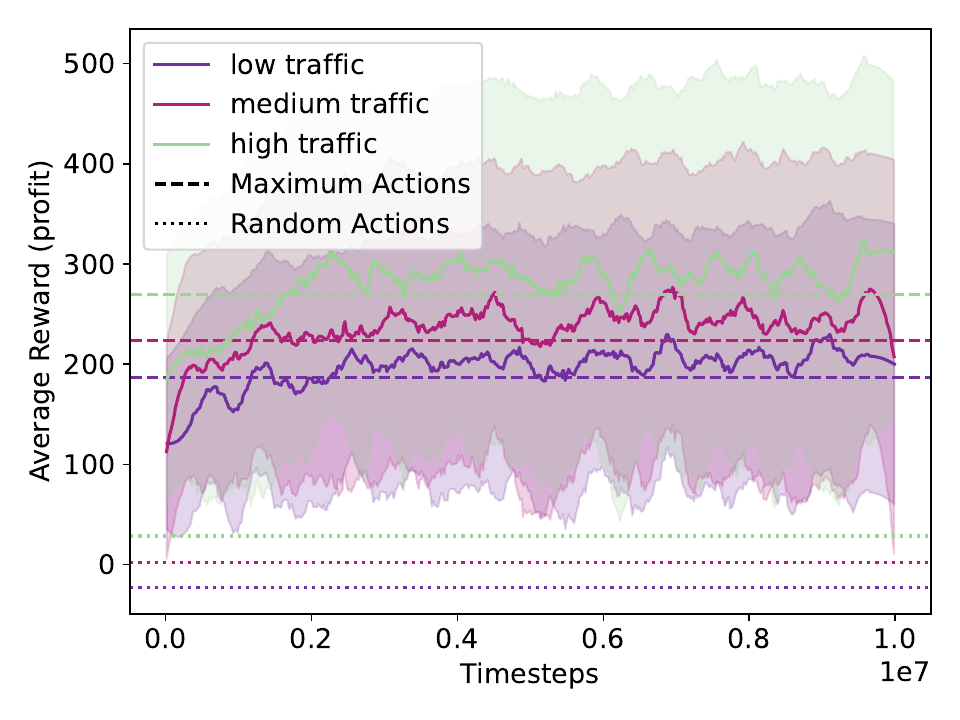}
    \caption{Workplace}
  \end{subfigure}
  \hfill
  \begin{subfigure}{0.245\textwidth}
    \centering
    \includegraphics[width=\textwidth]{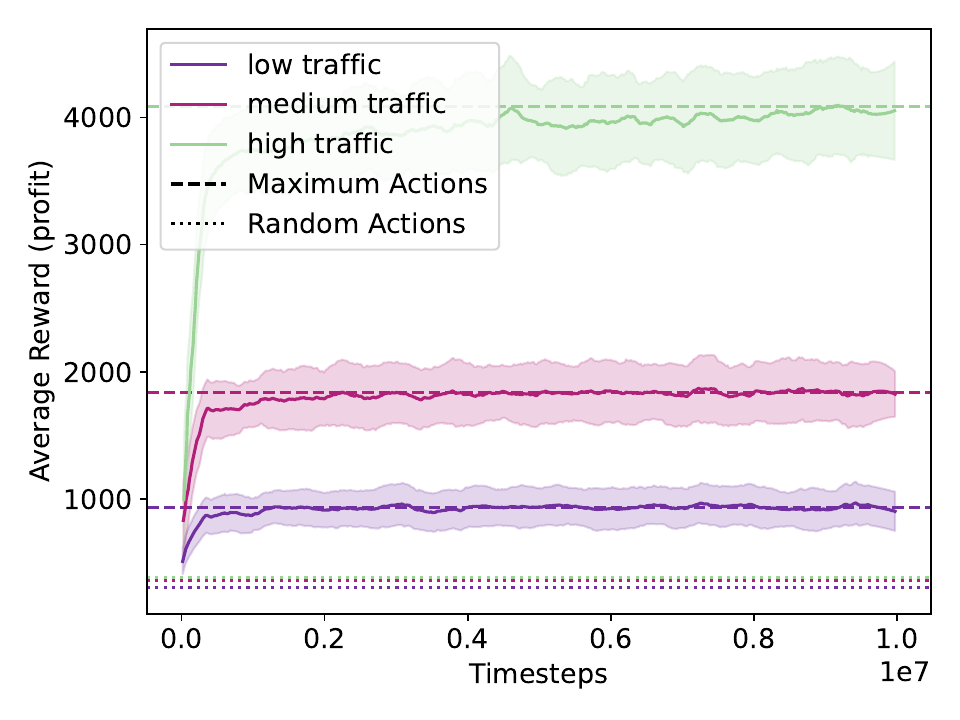}
    \caption{Highway}
  \end{subfigure}
  \caption{Results on our 4 bundled scenarios using EU cars and 8 AC (11.5kW) and 8 DC (150kW) chargers}
\end{figure}

\begin{figure}[ht]
  \centering
  \begin{subfigure}{0.245\textwidth}
    \centering
    \includegraphics[width=\textwidth]{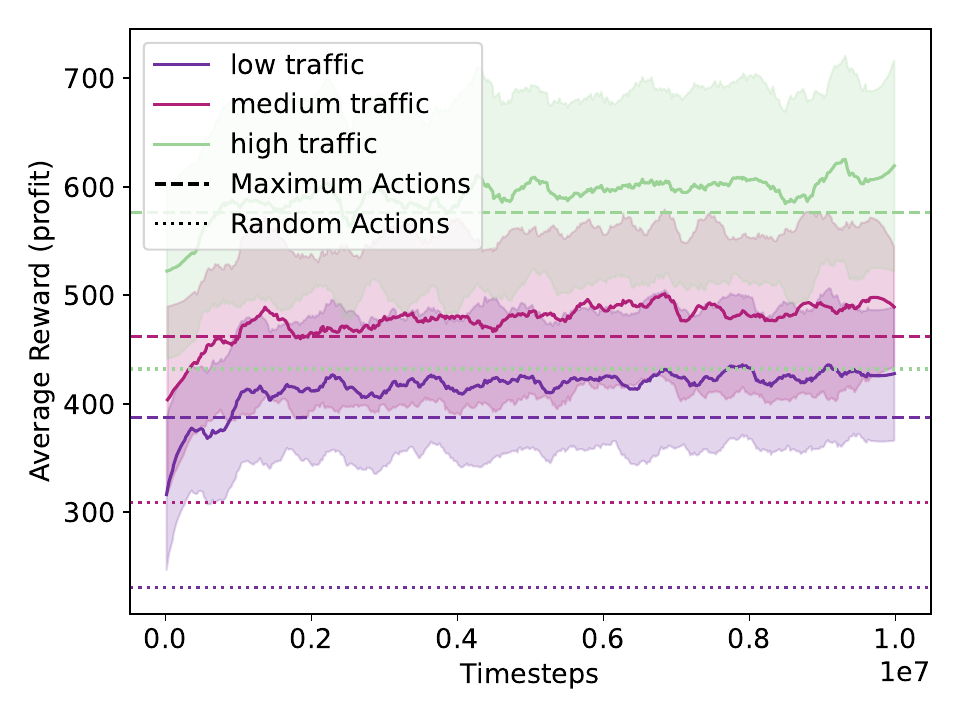}
    \caption{Residential}
  \end{subfigure}
  \hfill
  \begin{subfigure}{0.245\textwidth}
    \centering
    \includegraphics[width=\textwidth]{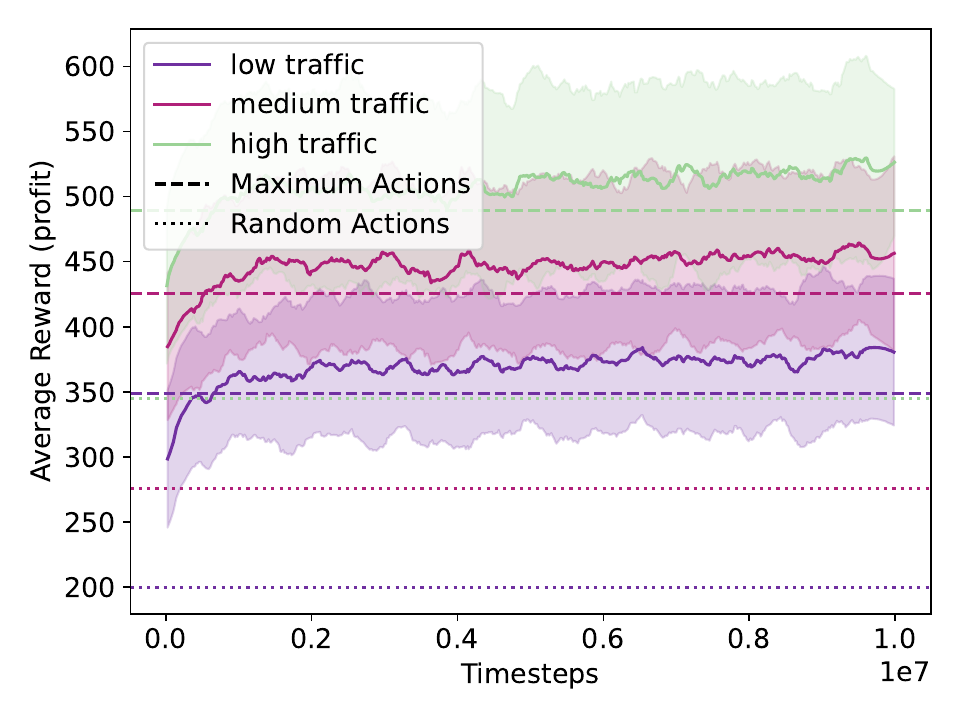}
    \caption{Shopping}
  \end{subfigure}
  \hfill
  \begin{subfigure}{0.245\textwidth}
    \centering
    \includegraphics[width=\textwidth]{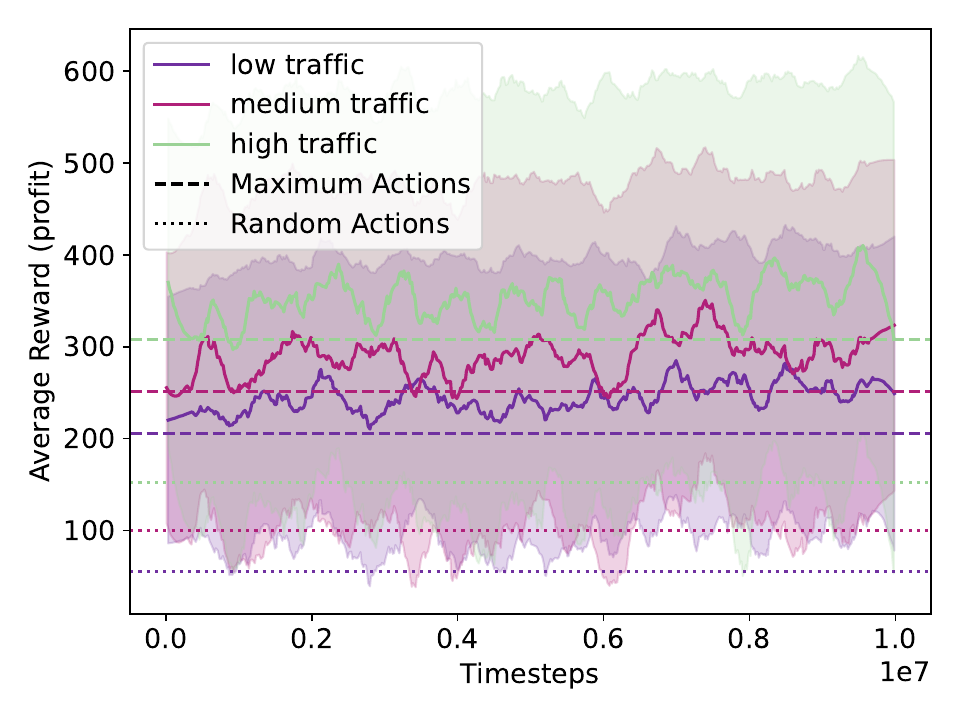}
    \caption{Workplace}
  \end{subfigure}
  \hfill
  \begin{subfigure}{0.245\textwidth}
    \centering
    \includegraphics[width=\textwidth]{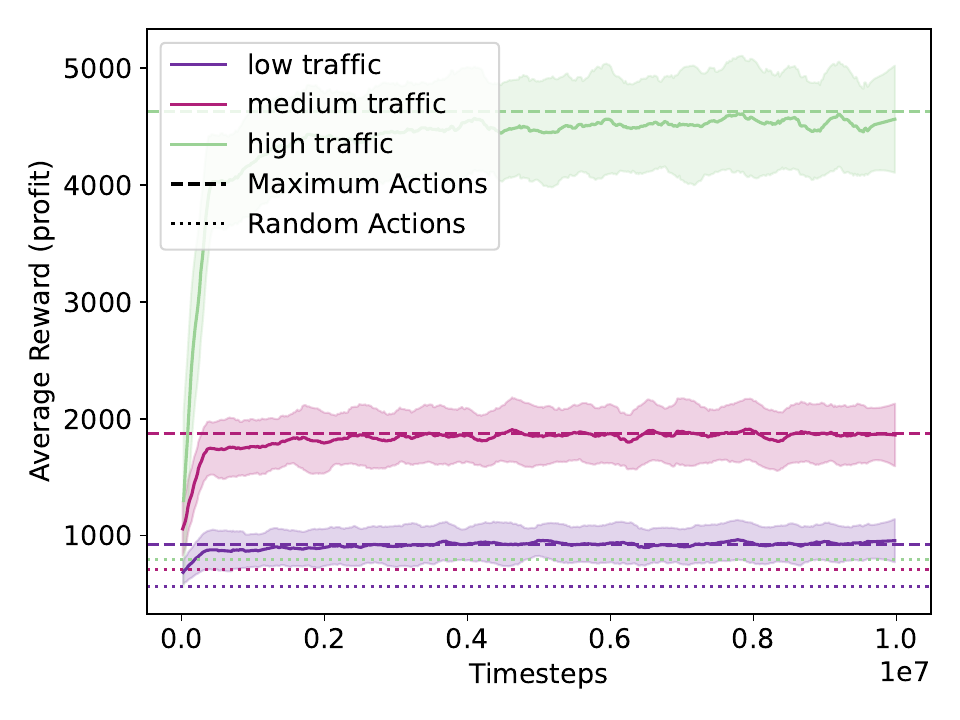}
    \caption{Highway}
  \end{subfigure}
  \caption{Results on our 4 bundled scenarios using EU cars and 16 DC (150kW) chargers}
\end{figure}

\end{document}